%% file: main.tex
\documentclass[12pt]{article} 
\usepackage{arxiv}

\usepackage[numbers,sort&compress]{natbib}

\input{packages}

\graphicspath{{fig/}}
\author{
  Arnd Koeppe\thanks{Corresponding author: \url{arnd.koeppe@kit.edu}} \\
  Karlsruhe Institute of Technology (KIT)\\
  Straße am Forum 7, 76131 Karlsruhe, Germany \\
  \And
  Franz Bamer \\
  RWTH Aachen University\\
  Eilfschornsteinstraße 18, 52062 Aachen, Germany \\
  \And
  Michael Selzer \\
  Karlsruhe Institute of Technology (KIT)\\
  Straße am Forum 7, 76131 Karlsruhe, Germany \\
  \And
  Britta Nestler \\
  Karlsruhe Institute of Technology (KIT)\\
  Straße am Forum 7, 76131 Karlsruhe, Germany \\
  \And
  Bernd Markert \\
  RWTH Aachen University\\
  Eilfschornsteinstraße 18, 52062 Aachen, Germany \\
}

\title{Explainable artificial intelligence for mechanics: physics-informing neural networks for constitutive models}

\renewcommand{\headeright}{}
\renewcommand{\undertitle}{}
\renewcommand{\shorttitle}{Explainable AI for mechanics}

\begin{document}

\maketitle


\begin{abstract}

(Artificial) neural networks have become increasingly popular in mechanics to accelerate computations with model order reduction techniques
and as universal models for a wide variety of materials.
However, the major disadvantage of neural networks remains: their numerous parameters are challenging to interpret and explain.
Thus, neural networks are often labeled as black boxes, and their results often elude human interpretation.
In mechanics, the new and active field of physics-informed neural networks attempts to mitigate this disadvantage by designing deep neural networks on the basis of mechanical knowledge.
By using this a priori knowledge, deeper and more complex neural networks became feasible, since the mechanical assumptions could be explained.
However, the internal reasoning and explanation of neural network parameters remain mysterious.

Complementary to the physics-informed approach, we propose a first step towards a physics-informing approach, which explains neural networks trained on mechanical data a posteriori.
This novel explainable artificial intelligence approach aims at elucidating the black box of neural networks and their high-dimensional representations.
Therein, the principal component analysis decorrelates the distributed representations in cell states of RNNs and allows the comparison to known and fundamental functions.
The novel approach is supported by a systematic hyperparameter search strategy that identifies the best neural network architectures and training parameters.
The findings of three case studies on fundamental constitutive models (hyperelasticity, elastoplasticity, and viscoelasticity) imply that the proposed strategy can help identify numerical and analytical closed-form solutions to characterize new materials.

\end{abstract}









\section{Introduction}

Data-driven models trained with deep learning algorithms have achieved tremendous successes in many research fields \cite{lecun2015deeplearning}.
As the archetypical deep learning model, (artificial) neural networks and their variants are powerful predictors exceptionally well-suited for spatio-temporal data, such as mechanical tensor fields \cite{koeppe2020intelligentnonlinearmeta}.
Each successive layer of a deep neural network learns to extract higher-level representations of the input and creates a data-driven model by supervised learning.
Due to the large variety of layers and cells (cf.\ \cite{goodfellow2016deeplearning}), neural networks are highly modular and successful for many applications \cite{lecun2015deeplearning}.

In mechanics, neural networks have been established as data-driven constitutive modeling and model order reduction techniques.
\citeauthor{ghaboussi1991knowledgebasedmodeling} \cite{ghaboussi1991knowledgebasedmodeling} proposed a unified constitutive model with shallow neural networks that learn from experimental data.
Further extensions reduced the required number of experimental samples \cite{ghaboussi1998autoprogressivetrainingneural}, adjusted the hidden layer dimensionality during training \cite{ghaboussi1998newnestedadaptive}, and approximated the stiffness matrix \cite{hashash2004numericalimplementationneural}.
\citeauthor{theocaris1995plasticityincludingbauschinger} used dense neural networks to model kinematic hardening \cite{theocaris1995plasticityincludingbauschinger} and identify parameters for the failure mode of anisotropic materials \cite{theocaris1997parameteridentificationproblem}.
\citeauthor{shin2000selflearningfiniteelement} \cite{shin2000selflearningfiniteelement} and \citeauthor{javadi2003neuralnetworkconstitutive} \cite{javadi2003neuralnetworkconstitutive,javadi2009intelligentfiniteelement,javadi2009intelligentfiniteelementa} proposed ``intelligent finite elements'', neural network constitutive models that were applied to soils under cyclic loading and tunneling processes.
Using context neurons in Recurrent Neural Networks (RNNs), \citeauthor{oeser2009modelingmaterialsfading} \cite{oeser2009modelingmaterialsfading},
\citeauthor{graf2012structuralanalysisfuzzy} \cite{graf2012structuralanalysisfuzzy}, and
\citeauthor{freitag2013materialdescriptionbased} \cite{freitag2013materialdescriptionbased}
modeled elastoplastic and viscoelastic materials with fuzzy parameters.
\citeauthor{bessa2017frameworkdatadrivenanalysis} \cite{bessa2017frameworkdatadrivenanalysis} proposed a data-driven analysis framework for materials with uncertain material parameters.
For cantilever beams, \citeauthor{sadeghi2017identificationnonlinearparameter} \cite{sadeghi2017identificationnonlinearparameter} used neural networks for nonlinear system identification and parameter estimation, while \citeauthor{koeppe2016modelreductionsubmodelling} \cite{koeppe2016modelreductionsubmodelling} used dense neural networks to predict the displacement response.
Extensions to continuum models resulted in linear ``intelligent meta element'' models of a cantilever beam \cite{koeppe2018intelligentmetaelementlinear}.
To bridge the gap between atomistic and continuum mechanics, \citeauthor{teichert2019machinelearningmaterials} \cite{teichert2019machinelearningmaterials} trained integrable deep neural networks on atomistic scale models and successively approximated free energy functions.
\citeauthor{stoffel2018artificialneuralnetworks} \cite{stoffel2018artificialneuralnetworks} used dense neural networks to fit material data from high-velocity shock-wave tube experiments.
\citeauthor{heider2020invarianceinformedgraphbaseddeep} \cite{heider2020invarianceinformedgraphbaseddeep} investigated the frame invariance for graph-based neural networks predicting anisotropic elastoplastic material behavior.
\citeauthor{huang2020machinelearningbased} \cite{huang2020machinelearningbased} combined the proper orthogonal decomposition, manual history variables, and dense neural networks for hyperelasticity and plasticity.

Despite the efforts to combine artificial intelligence and mechanics, the main disadvantage of neural networks remains: the learned parameters in black-box neural networks are challenging to interpret and explain \cite{breiman2001statisticalmodelingtwo}.
Their high-level representations in deeper layers often elude human interpretation: what the neural network understood and how the individual parameter values can be explained remains incomprehensible.
Coupled with insufficient data and limited computation capacities for past efforts, engineers and scientists mistrusted neural networks in favor of simpler models, whose fewer parameters could be easily interpreted and explained.
Occam's razor, the well-known problem-solving principle by William of~Ockham (ca.\ \numrange{1287}{1347}), became an almost dogmatic imperative: the simpler model with fewer parameters must be chosen if two models are equally accurate.
However, with the advent of deep learning, the concept of simple models has become ambiguous. The shared representations common in dense neural networks require potentially infinite numbers of parameters to model arbitrary functions \cite{hornik1989multilayerfeedforwardnetworks}.
To handle the large dimensionality of mechanical spatio-temporal data, extensive parameter sharing, utilized by recursions \cite{freitag2013materialdescriptionbased,freitag2017recurrentneuralnetworks,koeppe2018efficientnumericalmodeling,koeppe2019efficientmontecarlo,koeppe2017neuralnetworkrepresentation} and convolutions \cite{koeppe2020intelligentnonlinearmeta,koeppe2020mechanikkuenstlicheintelligenz,wu2020datadrivenreducedorder}, introduces assumptions based on prior knowledge and user-defined parameters, i.e., hyperparameters, to reduce the number of trainable parameters.

By deriving the prior assumptions on the deep learning algorithm from mechanical knowledge, the recent trend in computational mechanics, enhanced by neural networks, aims towards physics-informed neural networks.
\citeauthor{lagaris1998artificialneuralnetworks} \cite{lagaris1998artificialneuralnetworks} and
\citeauthor{aarts2001neuralnetworkmethod} \cite{aarts2001neuralnetworkmethod} imposed boundary and initial conditions on dense neural networks, which were trained to solve unconstrained partial differential equations by differentiating the neural network graphs.
In \cite{ramuhalli2005finiteelementneuralnetworks}, \citeauthor{ramuhalli2005finiteelementneuralnetworks} designed their neural network by embedding a finite element model, which used the neural network's hidden layer to predict stiffness components and the output layer to predict the external force.
\citeauthor{baymani2010artificialneuralnetworks} \cite{baymani2010artificialneuralnetworks} split the Stokes problem into three Poisson problems, approximated by three neural networks with superimposed boundary condition terms.
In \cite{rudd2014constrainedbackpropagationapproach}, \citeauthor{rudd2014constrainedbackpropagationapproach} used a constrained backpropagation approach to ensure that the boundary conditions are satisfied at each training step.
The physics-informed neural network approach, as proposed by \citeauthor{raissi2019physicsinformedneuralnetworks} \cite{raissi2019physicsinformedneuralnetworks},
employed two neural networks that contributed to the objective function. One physics-informed neural network mimicked the balance equation, while the second neural network approximated the potential function and pressure field.
\citeauthor{yang2019adversarialuncertaintyquantification} \cite{yang2019adversarialuncertaintyquantification} employed adversarial training to improve the performance of physics-informed neural networks.
In \cite{kissas2020machinelearningcardiovascular}, \citeauthor{kissas2020machinelearningcardiovascular} developed physics-informed neural networks to model arterial blood pressure with magnetic resonance imaging data.
Using the FEM to inspire a novel neural network architecture, \citeauthor{koeppe2020intelligentnonlinearmeta} \cite{koeppe2020intelligentnonlinearmeta} developed a deep convolutional recurrent neural network architecture that maps Dirichlet boundary constraints to force response and discretized field quantities in intelligent meta elements.
\citeauthor{yao2020feanetphysicsguideddatadriven} \cite{yao2020feanetphysicsguideddatadriven} used physics-informed deep convolutional neural networks to predict mechanical responses.
Thus, using physics-informed approaches, neural networks with more parameters became feasible, whose architecture could be designed and explained using mechanical knowledge.
However, the internal reasoning of the neural networks remained challenging to understand.

As opposed to the physics-informed approach, this work constitutes the first step towards a radically different search approach to the aforementioned design approach.
Inspired by the explainable Artificial Intelligence (AI) research field \cite{montavon2019layerwiserelevancepropagation,samek2019explainableaiinterpreting,alber2018innvestigateneuralnetworks,montavon2018methodsinterpretingunderstanding,bach2015pixelwiseexplanationsnonlinear,arras2017explainingrecurrentneural}, this novel approach elucidates the black box of neural networks to provide \emph{physics-informing} neural networks that complement the existing physics-informed neural networks.
As a result, more powerful neural networks may be employed with confidence, since both the architectural design and the learned parameters can be explained.

This work has the objective to efficiently and systematically search neural network architectures for fundamental one-dimensional constitutive models and explain the trained neural networks.
For this explanation, we propose a novel explainable AI method that uses the principal component analysis to interpret the cell states in RNNs.
For the search, we define a wide hyperparameter search domain, implement an efficient search algorithm, and apply it to hyperelastic, viscoelastic, and elastoplastic constitutive models.
After the search, the best neural network architectures are trained to high accuracy, and their generalization capabilities are tested.
Finally, the developed explainable AI approach compares the temporal behavior of RNN cell states to known solutions to the fundamental physical problem, thereby demonstrating that neural networks without prior physical assumptions have the potential to inform mechanical research.

To the best of the authors' knowledge, this work constitutes the first article proposing a dedicated explainable AI strategy for neural networks in mechanics, as well as a novel approach within the general field of explainable AI.
Unique to RNNs, which are popular in mechanics \cite{freitag2011predictiontimedependentstructural,freitag2017recurrentneuralnetworks,cao2016hybridrnngpodsurrogate,koeppe2018efficientnumericalmodeling,koeppe2019efficientmontecarlo,wu2020recurrentneuralnetworkaccelerated,graf2010recurrentneuralnetworks}, this work complements popular strategies for classification problems that investigate kernels in convolutional neural networks or investigate the data flow through neural networks, such as layer-wise relevance propagation \cite{bach2015pixelwiseexplanationsnonlinear,montavon2019layerwiserelevancepropagation,samek2019explainableaiinterpreting}.
Moreover, our new approach complements unsupervised approaches to find parsimonious and interpretable models for hyperelastic material behavior \cite{flaschel2021unsuperviseddiscoveryinterpretable}, which are one of the most recent representatives of the principle of simplicity.
Finally, the systematic hyperparameter search strategy offers an alternative strategy to self-designing neural networks, which have successfully modeled anisotropic and elastoplastic constitutive behavior \cite{fuchs2021dnn2hyperparameterreinforcement,heider2021offlinemultiscaleunsaturated}.

In \cref{sec:theory}, we briefly review the necessary preliminaries for RNN training as the foundation for the following sections.
\Cref{sec:methods} details the systematic hyperparameter search strategy and explainable AI approach for data-driven constitutive models.
For fundamental constitutive models, \cref{sec:results} demonstrates the developed approaches in three case studies.
Finally, \cref{sec:conclusion} summarizes and concludes this paper.

\section{Preliminaries}
\label{sec:theory}

\subsection{Training feedforward and recurrent neural networks}
\label{ss:theory-nn}

Dense feedforward neural networks are the fundamental building block of neural networks.
They represent nonlinear parametric mappings $\FUN{NN}$ from inputs $\ARR{x}$ to predictions $\hat{\ARR{y}}$ with $L$ consecutive layers and trainable parameters $\ARR{\theta}$:
\begin{equation}
  \FUN{NN}: \quad \ARR{x} ; \ARR{\theta} \longmapsto \hat{\ARR{y}}
  \quad \text{with} \quad \ARR{\theta} = \brk[c]2{
  \brk[r]1{\ARR{W}^{(1)}, \ARR{b}^{(1)}}
  \dots
  \brk[r]1{\ARR{W}^{(L)}, \ARR{b}^{(L)}}
  }
  \;,
\end{equation}
For dense feedforward neural networks, the parameters $\ARR{\theta}$ include the layer weights \smash{$\ARR{W}^{(l)}$} and biases \smash{$\ARR{b}^{(l)}$}.
Each layer $l$ applies a linear transformation to the layer inputs \smash{$\ARR{x}^{(l)}$}, before a nonlinear activation function \smash{$\FUN{f}^{(l)}$} returns the layer activations \smash{$\ARR{a}^{(l)}$}
\begin{equation}
  \ARR{a}^{(l)} = \mathrm{}\FUN{f}^{(l)}(\ARR{z}^{(l)})
  \quad \text{with} \quad \ARR{z}^{(l)} = \ARR{W}^{(l)} \ARR{x}^{(l)} + \ARR{b}^{(l)}
  \quad \forall l = 1 \dots L \;.
\end{equation}
Assembling and vectorizing these consecutive layers from \smash{$\ARR{x} \equiv \ARR{a}^{(0)}$} to \smash{$\hat{\ARR{y}} \equiv \ARR{a}^{(L)}$} enables fast computations on Graphics Processing Units (GPUs) and Tensor Processing Units (TPUs).

Using supervised learning, neural network training identifies optimal parameters $\ARR{\theta}^{\text{opt}}$, which minimize the difference between desired outputs and neural network predictions.
From a wide variety of input-target samples $\brk[r]1{\ARR{x}, \ARR{y}}$, the task of predicting physical function values $\hat{\ARR{y}} \approx \ARR{y}$ can be reformulated as a machine learning problem, where the function value represents the expectation of the data probability distribution $\PROB{p}^{\text{data}}(\ARR{y} \mid \ARR{x})$:
\begin{equation}
  \hat{\ARR{y}}(\ARR{x}) \approx \STAT{E}\brk[s]{\PROB{p}^{\text{data}}(\ARR{y} \mid \ARR{x})}
  \;.
\end{equation}

Assuming a single-peak distribution, e.g., a normal distribution $\PROB{p}^{\text{data}}(\ARR{y} \mid \ARR{x}) \approx \DIST{N}(\ARR{y} \mid \ARR{x})$, the maximum likelihood estimation principle (cf.\ \citeauthor{goodfellow2016deeplearning} \cite{goodfellow2016deeplearning}) yields a compatible loss function $\FNCTL{L}(\ARR{y}, \hat{\ARR{y}})$ for the prediction of physical values, such as the Mean Squared Error (MSE). 
Using this loss function to compute scalar errors $\epsilon$, the contributions of each parameter to the error can be backpropagated to compute gradients and train the neural network with gradient descent.
To ensure generalization to unknown samples, the datasets used are split (usually randomly) into training, validation, and test sets. 
Gradient descent with backpropagation is only performed on the training set, while the validation set safeguards against overfitting of the parameters $\ARR{\theta}$.
The test set safeguards against overfitting of the chosen hyperparameter values $\ARR{\gamma}$, which include, e.g., the dimensionalities of each layer $\operatorname{dim}(\ARR{a}^{(l)})$ and the number of layers $L$.

Recurrent neural networks \cite{rumelhart1986learningrepresentationsbackpropagating} introduce time-delayed recursions to feedforward neural networks \cite{goodfellow2016deeplearning}.
Thus, RNNs use parameter sharing to reuse parameters efficiently for sequential data-driven problems.
Such sequential data $\ARR{x}_t$ may be sorted according to the rate-dependent real time or a pseudo time.
One straightforward implementation of an RNN cell introduces a single tensor-valued recursion as given by
\begin{align}
  \label{eq:nn-rnn0}
  \bar{\ARR{x}}_{t+1} &= \begin{bmatrix}\ARR{x}_{t+1} & \ARR{a}_{t}\end{bmatrix}^{\mathsf{T}} 
  \;,\\
  \label{eq:nn-rnn1}
  \ARR{c}_{t+1} &= \FUN{f}(\ARR{W} \bar{\ARR{x}}_{t+1} + \ARR{b}) \;, 
  \\
  \label{eq:nn-rnn3}
  \ARR{a}_{t+1} &= \ARR{W}^{\mathrm{o}} \ARR{c}_{t+1} + \ARR{b}^{\mathrm{o}} \;. 
\end{align}

\begin{figure}[htbp]
  \centering
  \includetikz{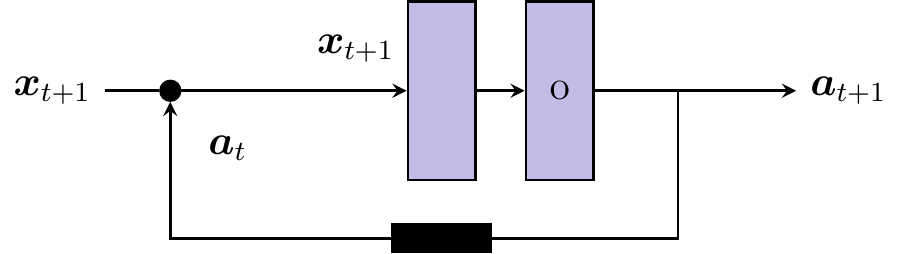}
  \caption{Visualization of a recurrent cell. The recurrent cell consists of two feedforward layers (light violet squares). A time-delayed recursion (black square) loops the activation of the first or last layer back as additional input to the cell.}
  \label{fig:nn-recurrent-cell}
\end{figure}

In \cref{fig:nn-recurrent-cell}, the recurrent cell is depicted as a graph of two conventional dense layers and a time-delayed recursion of previous activation $\ARR{a}_{t}$.
Subsequently embedded into a larger sequence tensor $\ARR{X} = \brk[s]1{\ARR{x}_1 \dots \ARR{x}_T}$, RNNs can be trained with backpropagation through time \cite{rumelhart1986learningrepresentationsbackpropagating}.

\begin{figure}[htbp]
  \centering
  \includetikz{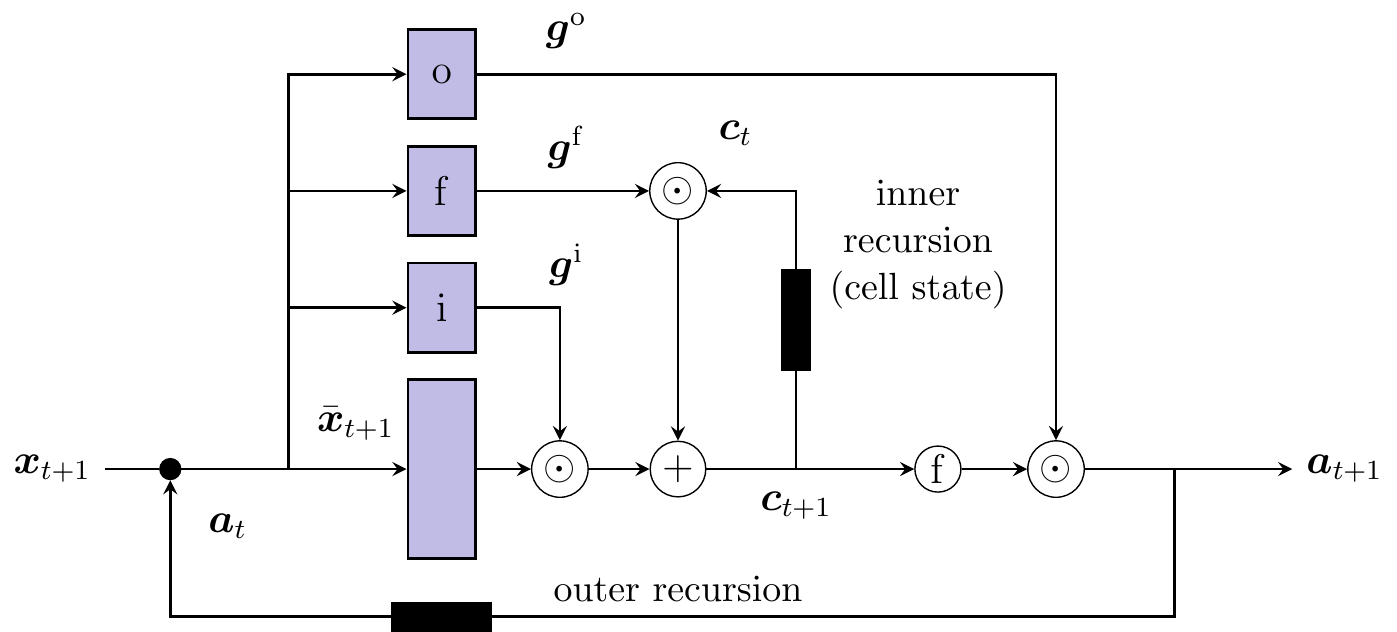}
  \caption{Visualization of an LSTM cell, which consists of multiple feedforward layers (light violet). Three gates, \smash{$\ARR{g}^{\mathrm{i}}$}, \smash{$\ARR{g}^{\mathrm{f}}$}, and \smash{$\ARR{g}^{\mathrm{o}}$}, control the data flow into the cell, within the cell, and out of the cell. Two recursions (black rectangles) ensure stable gradients across many time steps.}
  \label{fig:nn-lstm-cell}
\end{figure}

However, simple RNNs (\cref{eq:nn-rnn0,eq:nn-rnn1,eq:nn-rnn3}) suffer from unstable gradients and fading memory \cite{hochreiter1997longshorttermmemory,greff2015lstMSEarchspace}, which motivated the development of RNNs with gates. 
These gating layers control the data flow, thereby stabilizing the gradients and mitigating the fading memory.
The Long Short-Term Memory (LSTM) cell (\cref{fig:nn-lstm-cell}) \cite{hochreiter1997longshorttermmemory,gers2000learningforgetcontinual} and the Gated Recurrent Unit (GRU) \cite{cho2014learningphraserepresentations} are the most common gated RNNs and demonstrated comparable performances \cite{chung2014empiricalevaluationgated}.
As in \cref{eq:nn-rnn0}, the previous activation $\ARR{a}_t$ is concatenated with the input $\ARR{x}_{t+1}$:
\begin{align}
  \label{eq:nn-lstm-outer-loop}
  \bar{\ARR{x}}_{t+1} &= \begin{bmatrix}\ARR{x}_{t+1} & \ARR{a}_{t}\end{bmatrix}^{\mathsf{T}} \;,\\
  \tilde{\ARR{x}}_{t+1} &= \bar{\ARR{x}}_{t+1}  
    \quad \text{or} \quad \tilde{\ARR{x}}_{t+1} = \begin{bmatrix}\bar{\ARR{x}}_{t+1} & \ARR{c}_{t}\end{bmatrix}^{\mathsf{T}}
  \;.
\end{align}

The gates process the adjusted input $\tilde{\ARR{x}}_{t+1}$ using dense layers, $\brk[r]1{\ARR{W}^{\mathrm{i}}, \ARR{b}^{\mathrm{i}}}$, $\brk[r]1{\ARR{W}^{\mathrm{f}}, \ARR{b}^{\mathrm{f}}}$, and $\brk[r]1{\ARR{W}^{\mathrm{o}}, \ARR{b}^{\mathrm{o}}}$, and apply a sigmoid activation, which yields
\begin{align}
  \label{eq:nn-lstm-input-gate}
  \ARR{g}^{\mathrm{i}} &= \mathrm{sig}(\ARR{W}^{\mathrm{i}} \tilde{\ARR{x}}_{t+1} + \ARR{b}^{\mathrm{i}})\;,\\
  \label{eq:nn-lstm-forget-gate}
  \ARR{g}^{\mathrm{f}} &= \mathrm{sig}(\ARR{W}^{\mathrm{f}} \tilde{\ARR{x}}_{t+1} + \ARR{b}^{\mathrm{f}})\;,\\
  \label{eq:nn-lstm-output-gate}
  \ARR{g}^{\mathrm{o}} &= \mathrm{sig}(\ARR{W}^{\mathrm{o}} \tilde{\ARR{x}}_{t+1} + \ARR{b}^{\mathrm{o}})\;.
\end{align}
Therein, $\ARR{g}^{\mathrm{i}}$ is the input gate activation, $\ARR{g}^{\mathrm{f}}$ the forget gate activation, and $\ARR{g}^{\mathrm{o}}$ the output gate activation, whose coefficient values are bounded between zero and one.
Similar to Boolean masks, these gates control the data flow in the LSTM cell by
\begin{align}
  \label{eq:nn-lstm-input-flow}
  \ARR{i}_{t+1} &= \ARR{g}^{\mathrm{i}} \odot \FUN{f}(\ARR{W} \bar{\ARR{x}}_{t+1} + \ARR{b}) \;,\\
  \label{eq:nn-lstm-cell-flow}
  \ARR{c}_{t+1} &= \ARR{g}^{\mathrm{f}} \odot \ARR{c}_{t} + \ARR{i}_{t+1} \;,\\
  \label{eq:nn-lstm-flow-out}
  \ARR{a}_{t+1} &= \ARR{g}^{\mathrm{o}} \odot \FUN{f}(\ARR{c}_{t+1}) \;,
\end{align}
where the element-wise product $\odot$ allows the gate tensors $\ARR{g}^{\mathrm{i}}$, $\ARR{g}^{\mathrm{f}}$, and $\ARR{g}^{\mathrm{o}}$ to control the incoming data flow into the cell $\ARR{c}_{t+1}$, to forget information from selected entries, and to output selected information as $\ARR{a}_{t+1}$.
Despite being the foundation for the RNN reasoning and long-term memory, the cell states are often regarded as black boxes.

\subsection{A selection of fundamental constitutive models}
\label{ss:theory-constitutive-models}

The systematic hyperparameter search and explainable AI approach will be demonstrated on three representative constitutive models, briefly reviewed in the following.
These well-known models describe material effects, such as finite, lasting, and rate-dependent deformations.
Since the ground truths for these one-dimensional models are known, the resulting neural network architectures from the hyperparameter search can be evaluated.
Furthermore, for history-dependent material behaviors, the explainable AI approach can investigate the RNNs that best describe the constitutive models.

\begin{figure}[htbp]
  \centering
  \subcaptionbox{
    Rheological model
    \label{fig:method-one-dimensional-hyperelasticity-rheological}}{
    \def\svgscale{0.5}
    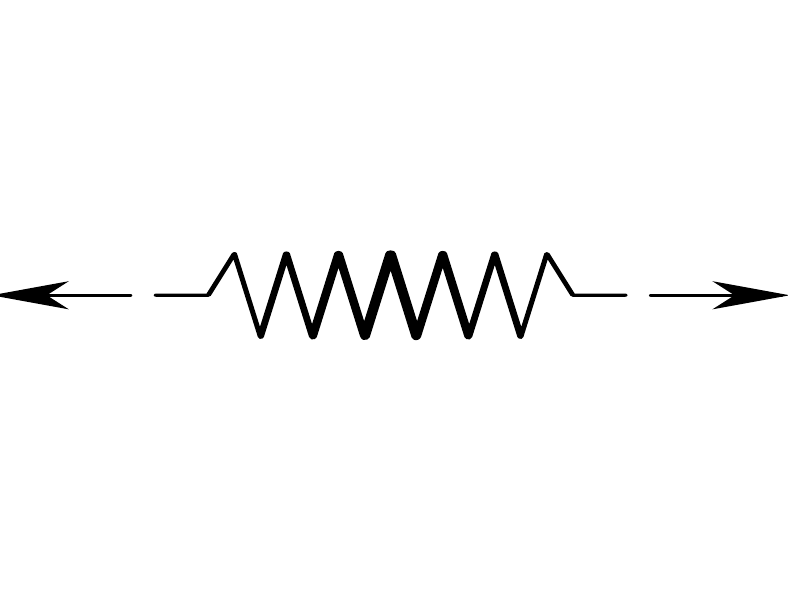
  }%
  \hspace{2.5cm}%
  \subcaptionbox{
    Example cycle
    \label{fig:method-one-dimensional-hyperelasticity-cycle}}{
      \includetikz{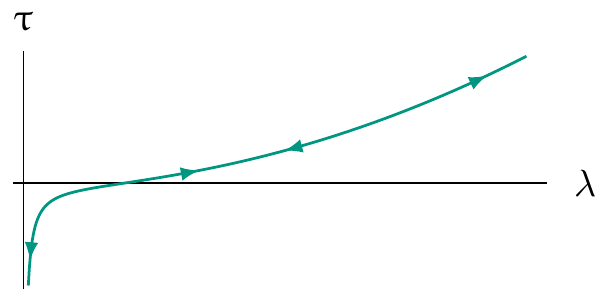}
  }
  \caption{One-dimensional hyperelastic material behavior.}
  \label{fig:method-one-dimensional-hyperelasticity}
\end{figure}

First, a hyperelastic constitutive model challenges neural networks to model finite strains and deal with the singular behavior of stretches that approach zero.
The one-dimensional Neo-Hooke model (cf. \citeauthor{holzapfel2000nonlinearsolidmechanics} \cite{holzapfel2000nonlinearsolidmechanics}) is regarded as one of the most straightforward nonlinear elastic material models.
Therein, a single parameter $\mu$ describes the Cauchy stress $\uptau$ as a function of the stretch $\uplambda$:
\begin{equation}
  \uptau = \mu \brk[r]*{\uplambda^2 - \frac{1}{\uplambda}} \;.
\end{equation}
Thus, the Neo-Hooke model offers the highest possible contrast to distributed neural network representations.
In \cref{fig:method-one-dimensional-hyperelasticity-rheological}, the corresponding rheological model (\cref{fig:method-one-dimensional-hyperelasticity-rheological}) and a loading and unloading cycle (\cref{fig:method-one-dimensional-hyperelasticity-cycle}) are depicted. 
The neural network approximation is challenged by stretch values near the origin, where the stress response exhibits singular behavior.
 
\begin{figure}[htbp]
  \centering
  \subcaptionbox{
    Rheological model
    \label{fig:method-one-dimensional-viscoelasticity-rheological}}{
    \def\svgscale{0.85}
    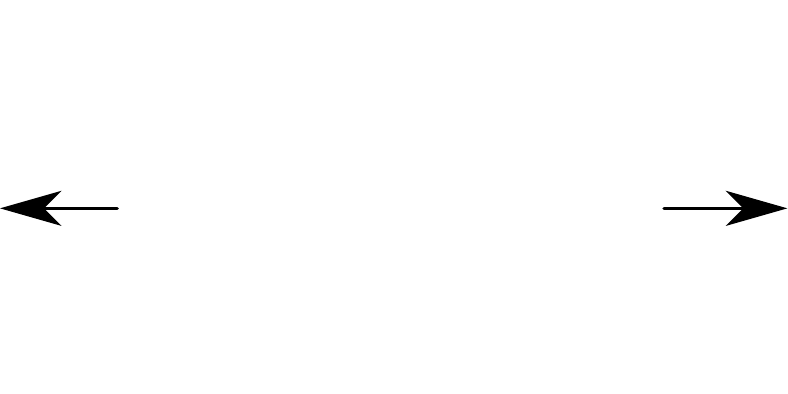
  }%
  \hfill%
  \subcaptionbox{
    Cyclic loading
    \label{fig:method-one-dimensional-viscoelasticity-cycle}}{
      \includetikz{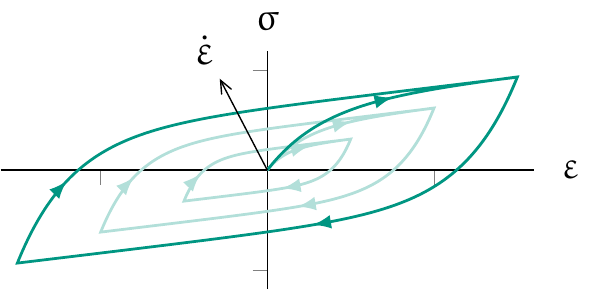}
  }\\
  \subcaptionbox{
    Stress relaxation
    \label{fig:method-one-dimensional-viscoelasticity-relax}}{
      \includetikz{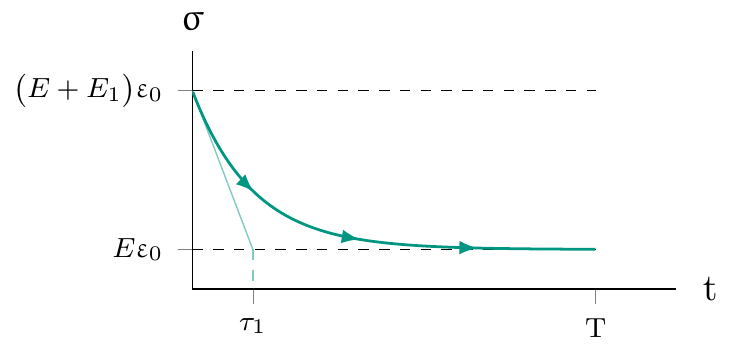}
  }%
  \hfill%
  \subcaptionbox{
    Creep retardation
    \label{fig:method-one-dimensional-viscoelasticity-creep}}{
      \includetikz{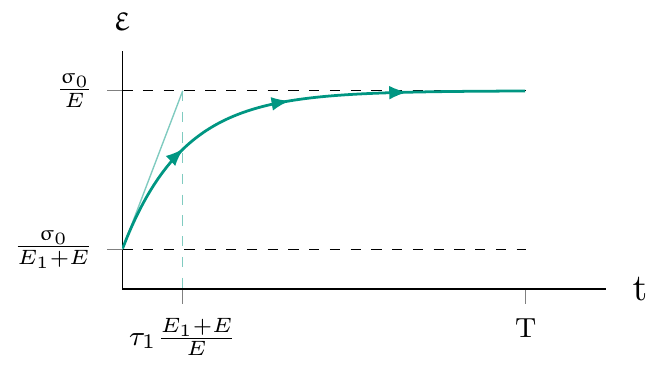}
  }
  \caption{One-dimensional viscoelastic material behavior. (a) Poynting-Thomson rheological model. (b) Cyclic loading hysteresis for constant strain rates $\dot{\upvarepsilon}$. (c) Stress relaxation response to a unit-sized step $\upvarepsilon(t) = \upvarepsilon_0 \operatorname{H}(t)$. (d) Creep retardation response to a unit-sized step $\upsigma(t) = \upsigma_0 \operatorname{H}(t)$.}
  \label{fig:method-one-dimensional-viscoelasticity}
\end{figure}

Second, viscoelasticity represents the archetype of fading-memory material behavior \cite{truesdell2004nonlinearfieldtheories}, which poses a short-term temporal regression problem to the neural networks.
\Cref{fig:method-one-dimensional-viscoelasticity-rheological} introduces the Poynting-Thomson or standard linear viscoelastic solid model as an example for rate-dependent inelastic material behavior.
As the elementary example for generalized Maxwell models, which use multiple parallel Maxwell branches to cover more decades in the frequency range, it exhibits the same principal relaxation and creep behavior \cite{markert2005porousmediaviscoelasticity}, as depicted in \cref{fig:method-one-dimensional-viscoelasticity-cycle}, \cref{fig:method-one-dimensional-viscoelasticity-relax}, and \cref{fig:method-one-dimensional-viscoelasticity-creep}.

The additive decomposition of the stress $\upsigma$ and the evolution equations for each Maxwell branch $q$, as reviewed, e.g., in the textbook of  \citeauthor{holzapfel2000nonlinearsolidmechanics} \cite{holzapfel2000nonlinearsolidmechanics}, yield
\begin{align}
  \label{eq:viscoelastic-evolution-branch}
  \dot{\upsigma}_q + \frac{\upsigma_q}{\tau_q} 
    &= E_q \dot{\upvarepsilon}
    \quad \forall\; q = 1 \dots Q
    \;, \quad \text{and}\\
  \label{eq:viscoelastic-stress-composition}
  \upsigma &= E \upvarepsilon + \sum\limits_q \upsigma_q
  \;.
\end{align}
For each branch $q$, $\upsigma_q$ represents the stress, $\tau_q$ the relaxation time, and $E_q$ the modulus. The strain $\upvarepsilon$ is shared by all Maxwell branches and the elastic spring with modulus $E$.
For $Q = 1$, \cref{eq:viscoelastic-evolution-branch,eq:viscoelastic-stress-composition} yield
\begin{equation}
  \label{eq:viscoelastic-poynting-thomson-ode}
  \upsigma + \tau_1 \dot{\upsigma}_q = E \upvarepsilon + \tau_1 \brk[r]{E + E_1} \dot{\upvarepsilon}
  \;.
\end{equation}
Many known excitations, e.g., unit steps $\upsigma(\mathrm{t}) = \upsigma_0 \operatorname{H}(\mathrm{t})$ or $\upvarepsilon(\mathrm{t}) = \upvarepsilon_0 \operatorname{H}(\mathrm{t})$ yield closed-form solutions described by the relaxation function $\operatorname{R}(\mathrm{t})$ and creep function $\operatorname{C}(\mathrm{t})$ \cite{simo1998computationalinelasticity}:
\begin{align}
  \label{eq:method-viscoelastic-relaxation-closed-form}
  \upsigma(\mathrm{t}) 
    &= \int\limits_{-\infty}^{\mathrm{t}} 
      \operatorname{R}(\mathrm{t} - \tilde{t}) \dot{\upvarepsilon}(\tilde{t}) \dd{\tilde{t}}
      \quad \text{with} \quad\\
    \operatorname{R}(\mathrm{t}) 
    &= E + E_1 \exp(-\frac{\mathrm{t}}{\tau_1}) \;,
    \\
  \label{eq:method-viscoelastic-creep-closed-form}
  \upvarepsilon(\mathrm{t}) 
    &= \int\limits_{-\infty}^{\mathrm{t}} 
      \operatorname{C}(\mathrm{t} - \tilde{t}) \dot{\upsigma}(\tilde{t}) \dd{\tilde{t}}
      \quad \text{with} \quad\\
    \operatorname{C}(\mathrm{t}) 
    &= \brk[s]*{
      \frac{1}{E + E_1} + \frac{E_1}{E (E + E_1)} 
      \brk[r]*{1 - \exp(-\frac{\mathrm{t} E}{\tau_1 (E + SEi)})}
    }
  \;.
\end{align}
For the general case of incremental excitations, numerical integration \cref{eq:viscoelastic-evolution-branch} with an implicit backward Euler scheme yields
\begin{equation}
  \label{eq:viscoelastic-poynting-thomson-ode-discretized}
  \begin{aligned}
  \upsigma_{\mathrm{t} + \Delta\mathrm{t}} = 
    \frac{1}{c_{\upsigma1}} 
    &
    \brk[s]2{
      c_{e1} \upvarepsilon_{\mathrm{t} + \Delta\mathrm{t}} 
      - c_{e0} \upvarepsilon_{\mathrm{t}} 
      + c_{\upsigma0} \upsigma_{\mathrm{t}}
    }
    \\
    \quad \text{with} \quad
    c_{\upsigma0} &= \frac{\tau_1}{\Delta\!\!\;\mathrm{t}} \;,\;\;
    c_{\upsigma1} = 1 + \frac{\tau_1}{\Delta\!\!\;\mathrm{t}} \;,
    \\
    c_{\upepsilon0} &= \frac{\tau_1}{\Delta\!\!\;\mathrm{t}} \brk[r]1{E + E_1} \;,\;\;
    c_{\upepsilon1} = E + \frac{\tau_1}{\Delta\!\!\;\mathrm{t}} \brk[r]1{E + E_1} \;.
  \end{aligned}
\end{equation}
Therein, \smash{$\upsigma_{\mathrm{t}}$} and \smash{$\upsigma_{\mathrm{t} + \Delta\mathrm{t}}$} represent the stresses while \smash{$\upvarepsilon_{\mathrm{t}}$} and \smash{$\upvarepsilon_{\mathrm{t} + \Delta\mathrm{t}}$} respectively represent the strains, respectively at the current and next increment. 
As a dissipative model, the dissipated energy can be described by the area surrounded by the hystereses, which depend on the strain rates (\cref{fig:method-one-dimensional-viscoelasticity-cycle}).
 
\begin{figure}[htbp]
  \centering
  \subcaptionbox{
    Rheological model
    \label{fig:method-one-dimensional-elastoplasticity-ip-rheological}}{
    \def\svgscale{0.65}
    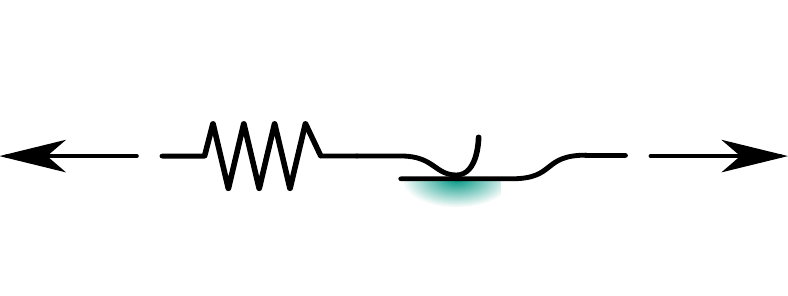
  }%
  \hfill%
  \subcaptionbox{
    Example cycle
    \label{fig:method-one-dimensional-elastoplasticity-cycle}}{
    \includetikz{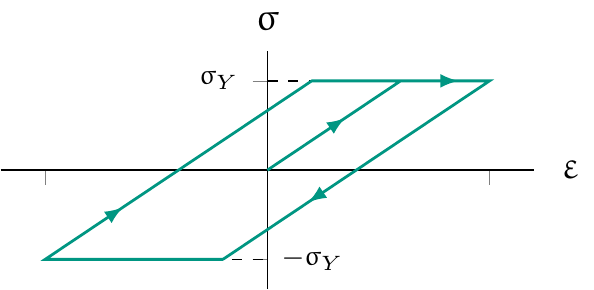}
  }
  \caption{One-dimensional Prandtl-Reuss elastoplastic material behavior with perfect plasticity.}
  \label{fig:method-one-dimensional-elastoplasticity}
\end{figure}

Finally, elastoplastic models exhibit path-dependent behavior with lasting deformations, i.e., long-term dependency behavior, and include potential discontinuities in the stress response, which need to be learned by the neural network.
Numerical implementations of one-dimensional Prandtl-Reuss plasticity (\cref{fig:method-one-dimensional-elastoplasticity-ip-rheological}) can be found, e.g., in \citeauthor{simo1998computationalinelasticity} \cite{simo1998computationalinelasticity} or 
\citeauthor{deborst2012nonlinearfiniteelement} \cite{deborst2012nonlinearfiniteelement}:
\begin{align}
  \label{eq:elastoplastic-yield-stress-strain-1d}
  \upsigma &= E \brk[r]1{\upvarepsilon - \upvarepsilon^{\mathrm{P}}} \;,\\
  \label{eq:elastoplastic-evolution-1d}
  \dot{\upvarepsilon}^{\mathrm{P}} &= \gamma \operatorname{sign}(\upsigma - q_{\upsigma})
    \quad \text{and} \quad \dot{q}_{\upvarepsilon} = \gamma
    \quad \text{and} \quad \dot{q}_{\upsigma} = \gamma H \operatorname{sign}(\upsigma - q_{\upsigma}) \;,\\
  \label{eq:elastoplastic-yield-1d}
  f_y\brk[r]1{\upsigma, q_{\upvarepsilon}, q_{\upsigma}} 
    &= \abs{\upsigma - q_{\upsigma}} 
    - \brk[r]1{\upsigma_Y + K q_{\upvarepsilon}}
  \;.
\end{align}
In \Cref{fig:method-one-dimensional-elastoplasticity-cycle}, the characteristic stress-strain curve of one-dimensional Prandtl-Reuss plasticity is shown.
 
\section{Explainable artificial intelligence for mechanics}
\label{sec:methods}

This section introduces data-driven constitutive models (\cref{ss:methods-intelligent-constitutive-models}), the systematic hyperparameter search strategy (\cref{ss:methods-hyperparameter-search}), and the novel explainable AI approach (\cref{ss:methods-explainable-ai}).

\subsection{Data-driven constitutive models for fundamental material behavior}
\label{ss:methods-intelligent-constitutive-models}

Data-driven constitutive models and intelligent finite elements were one of the first applications of neural networks within the FEM \cite{ghaboussi1991knowledgebasedmodeling,ghaboussi1998autoprogressivetrainingneural,ghaboussi1998newnestedadaptive,javadi2009intelligentfiniteelement}, since they leverage the flexibility of the FEM to the fullest.
For a strain-driven problem, data-driven constitutive models can be defined by
\begin{equation}
  \mathrm{IC}: \quad 
  \ARR{\upvarepsilon} \quad 
  \operatorname*{\longmapsto}^{\ARR{Q}}\quad
  \ARR{\upsigma}
  \;,
\end{equation}
where $\ARR{\upvarepsilon}$ represents the strain and $\ARR{\upsigma}$ represents the stress. 
The history variables are gathered in ${\ARR{Q}}$, which may include both algorithmic history variables, such as the plastic strain $\upvarepsilon^{\mathrm{P}}$, and recurrent neural network cell states $\ARR{C}$.
As in the conventional FEM, the unified interface of inputs, outputs, and history variables enables a straightforward substitution of different data-driven constitutive models.
For example, data-driven constitutive models, e.g., trained on experimental data \cite{stoffel2018artificialneuralnetworks}, can substitute analytically derived constitutive models with trivial implementation effort.
However, this flexibility massively increases the choice of conceivable neural network architectures and their defining hyperparameter configurations of data-driven constitutive models. 
Often, these hyperparameter configurations are chosen by the user or tuned using brute-force search algorithms.

\subsection{Systematic hyperparameter search}
\label{ss:methods-hyperparameter-search}

Systematic hyperparameter search strategies constitute elegant and efficient solutions to finding optimal architectures and hyperparameter configurations.
Hyperparameter search algorithms automatize the task of identifying advantageous neural network architectures and tuning hyperparameters $\ARR{\gamma}$ to achieve good performance.
Since neural networks are nonlinear function approximators with numerous parameters, the error surface is generally non-convex, high dimensional, potentially non-smooth, and noisy \cite{li2017hyperbandnovelbanditbased}.
Furthermore, hyperparameters are interdependent, and the effects on the model remain unclear and problem-specific \cite{li2017hyperbandnovelbanditbased}, which often popularized brute force search algorithms, such as grid search \cite{bergstra2012randoMSEarchhyperparameter}.

Grid search is the most fundamental search algorithm that seeks to cover the entire hyperparameter search domain.
All possible combinations of user-defined hyperparameter values are tested one by one, with a fixed step size. 
The number and intervals of tested hyperparameter configurations $\ARR{\Gamma}$ are set arbitrarily by the user, making the approach wasteful, inefficient, and infeasible for large numbers of hyperparameters.

Random search algorithms \cite{bergstra2012randoMSEarchhyperparameter} yield probabilistic approaches for larger numbers of hyperparameters.
With enough trials, random search statistically explores the entire search space by testing a random selection of all possible combinations with varying step sizes.
Thus, high-dimensional search spaces are explored faster, which makes random search a widely used search algorithm for hyperparameter-intensive neural networks \cite{bergstra2012randoMSEarchhyperparameter}. 
Unfortunately, the computational effort of random search remains significant.

Therefore, this work follows approaches that increase the efficiency of the random search algorithm, leveraging the observation that the final performances of neural networks can be approximately assessed after a few epochs $n$, i.e., iterations through the entire training dataset.
The Successive Halving algorithm \cite{jamieson2016nonstochasticbestarm}, for example, trains randomly chosen hyperparameter configurations \smash{$\ARR{\Gamma} = \brk[c]1{\ARR{\gamma}^{(1)}\dots\ARR{\gamma}^{(C)}}$} for $n$ epochs. After that, the lowest-performing configurations are discarded, while the best-performing configurations are trained further. 
Thus, the approach focuses the computational resources on promising hyperparameter configurations.

Unfortunately, the number of epochs to train before deciding on discarding low-performing configurations constitutes another hyperparameter, which depends on the mechanical problem.
Either Successive Halving can explore more configurations $C$ for fewer epochs $n$ or investigate fewer configurations $C$ for more epochs $n$ per decision.
\citeauthor{li2017hyperbandnovelbanditbased} \cite{li2017hyperbandnovelbanditbased} solved this exploration issue with the Hyperband search algorithm (\Cref{alg:Hyperband}).
By gathering groups of configurations in $h$ brackets, Successive Halving can be applied for different numbers of epochs $n_h$ per group.
The first bracket $\mathrm{s} = h$ maximizes exploration to test
many configurations $C$ with Successive Halving, identifying promising positions even for vast search domains and non-convex or non-smooth loss functions.
In the last bracket $\mathrm{s} = 0$, fewer configurations are tested for the full number of epochs, similar to a conventional random search.
This is advantageous if the search domain is narrow and the objective functions are convex and smooth.
Thus, Hyperband combines efficient exploration and investigation of promising architectures and does not introduce additional hyperparameters to be tuned.

In this work, we combine Hyperband with aggressive early stopping, which discards solutions that are unlikely to improve further by monitoring moving averages of the validation errors.
Independent of Hyperband, this additional logic further increases the efficiency of the search strategy for data-driven constitutive models.
Thus, this fully automatized strategy identifies optimal neural network architectures for the given training and validation data without user interaction.
The resulting hyperparameter configurations represent a `natural' choice for the given problem, which merits further investigations into why this specific neural network was chosen.

\begin{algorithm}[bp]
  \caption{Hyperband search \cite{li2017hyperbandnovelbanditbased}}
  \label{alg:Hyperband}
  \setlength{\abovedisplayskip}{0.5ex}
  \setlength{\belowdisplayskip}{-2ex}
  \setlength{\abovedisplayshortskip}{0.5ex}
  \setlength{\belowdisplayshortskip}{-2ex}
  \DontPrintSemicolon
  \SetKwInput{KwInput}{input}
  \SetKwInput{KwInit}{initialization}
  \KwInput{
    maximum epochs per configuration $N$, 
    keep quotient $\eta$}
  \nl\KwInit{
    number of brackets $h \coloneqq \floor*{\log_{\eta}(N)}$,\\ 
    total epochs for Hyperband $B \coloneqq (h + 1) N$}
  \nl\For{$\mathrm{s} = h \dots 0$}{
    \nl number of initial configurations in bracket
      \begin{algomathdisplay}
        C \coloneqq \ceil*{\frac{B}{N} \frac{\eta^{\mathrm{s}}}{\mathrm{s} + 1}}
      \end{algomathdisplay}\;
    \nl maximum epochs per configuration in current bracket
      \begin{algomathdisplay}
        n \coloneqq \frac{N}{\eta^{\mathrm{s}}}
      \end{algomathdisplay}\;
    \nl sample set of hyperparameter configurations
      \begin{algomathdisplay}
        \ARR{\Gamma} \coloneqq \brk[c]2{\ARR{\gamma}^{(1)}\dots\ARR{\gamma}^{(C)}} \quad \text{with} \quad \ARR{\gamma} \sim \PROB{p}^{\text{hyper}}
      \end{algomathdisplay}\;
    \nl\For{$\mathrm{h} = 0 \dots \mathrm{s}$}{
      \nl number of configurations in current Successive Halving step
        \begin{algomathdisplay}
          C_h \coloneqq \floor*{\frac{C}{\eta^{\mathrm{h}}}}
        \end{algomathdisplay}\;
      \nl epochs in current Successive Halving step
        \begin{algomathdisplay}
          n_h \coloneqq n \eta^{\mathrm{h}}
        \end{algomathdisplay}\;
      \nl train for $n_h$ epochs and compute performance
        \begin{algomathdisplay}
          \ARR{\epsilon} = \brk[c]*{\FNCTL{L}(\FUN{NN}_{\ARR{\gamma}}\brk[r]*{\ARR{X}^{\text{valid}}; \ARR{\theta}_{\ARR{\gamma}}(n_h)}, \ARR{Y}^{\text{valid}}) \quad \forall \ARR{\gamma} \in \ARR{\Gamma}}
        \end{algomathdisplay}\;
      \nl reduce configuration set by $\eta$ based on performance
        \begin{algomathdisplay}
          \ARR{\Gamma} = \operatorname{select\_best\_k\_configurations}\brk[r]*{\ARR{\Gamma}, \ARR{\epsilon}, \mathrm{k}=\floor*{\tfrac{C}{\eta}}}
        \end{algomathdisplay}\;
    }
  }
  \Return{\normalfont$\ARR{\gamma}^{\text{best}} = \operatorname{argmin}{\ARR{\epsilon}(\ARR{\gamma})} \quad \text{with} \quad \ARR{\gamma} \in \ARR{\Gamma}$}
\end{algorithm}

\subsection{The novel explainable artificial intelligence approach}
\label{ss:methods-explainable-ai}

If the problem allows, it is generally possible to identify and train an efficient neural network to achieve accurate results and generalize to unknown data with the systematic approach outlined above.
However, in the past, the proven approximation capabilities of neural networks \cite{hornik1989multilayerfeedforwardnetworks} were often shunned because the magical black-box-generated results could not be explained.
Given only finite dataset sizes, any machine learning algorithm may learn spurious or accidental correlations.
Since neural networks naturally develop distributed high-dimensional representations in deep layers, the `reasoning' of neural networks is notoriously difficult to verify.
Thus, a trained neural network may generalize but use wrong assumptions to predict correct outputs.
The motivation of explainable AI is to unmask such `Clever Hans' predictors%
\footnote{Apparently, the horse `Clever Hans' (\numrange{1895}{1916}) could count and calculate, but, in fact, interpreted the expressions and gestures of the audience to find the correct answers.} %
\cite{lapuschkin2019unmaskingcleverhans}.

Many explainable neural network approaches, such as Layer-wise Relevance Propagation (LRP) \cite{bach2015pixelwiseexplanationsnonlinear,montavon2018methodsinterpretingunderstanding,montavon2019layerwiserelevancepropagation,arras2017explainingrecurrentneural}, 
use the neural network graph to trace the activation back to its origin.
In particular, such approaches are attractive for classification, because they can explain individual class labels, i.e., from single binary values to multiple inputs. 
For positive binary values, the activations can be traced back straightforwardly, explaining why the neural network chose the class associated with the binary value.
However, multivariate regression problems are faced with a different problem:
since the outputs are continuous, the neural network `reasoning' must be interpretable over the full output ranges, including the origin.
In particular, in balance equations, zero-valued residuals are at least as important as non-zero residuals and follow the same physical governing equations.
Thus, for mechanical regression tasks, different explainable AI methods are necessary, which focus on the evolution of the high-dimensional representations.

As the first explainable AI approach using mechanical domain knowledge, our proposed approach focuses on the temporal evolution of mechanical quantities.
Since time-variant problems in mechanics are often modeled by training RNNs on time-variant and path-dependent mechanical data \cite{koeppe2019efficientmontecarlo,freitag2011predictiontimedependentstructural,cao2016hybridrnngpodsurrogate,wu2020recurrentneuralnetworkaccelerated}, we propose to use the Principal Component Analysis (PCA) to investigate recurrent cell states, e.g., in LSTM and GRU cells.
Therefore, we interpret the time-variant cell states as statistical variables and use the PCA to identify the major variance directions in the distributed representations.
With the original evolution equations and history variables known, major principal components can be compared with the known temporal evolution.
If the cell states resemble the mechanical evolution equations of the algorithmic history variables, the neural network correctly understood the fundamental mechanical problem.
For future materials, neural networks can thus be trained on new material test data, and the material can be possibly characterized by comparing the cell state principal components to known fundamental evolution equations.

\begin{figure}[htbp]
  \centering
  \begin{subfigure}[c]{0.3\textwidth}
    \centering
    \includetikz{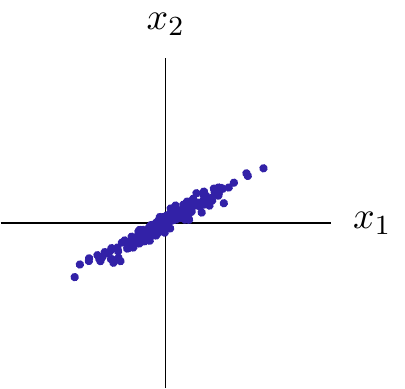}
    \subcaption{Original data \label{fig:method-feature-pca-rotated}}
  \end{subfigure}%
  \hspace{1cm}%
  \begin{subfigure}[c]{0.3\textwidth}
    \centering
    \includetikz{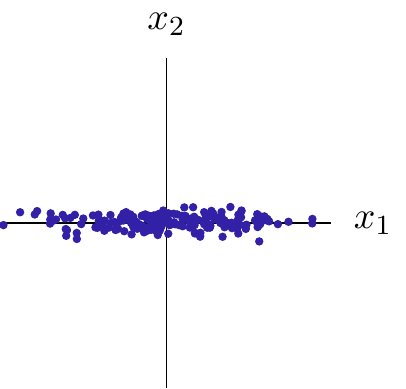}
    \subcaption{Decoupled data\label{fig:method-feature-pca-decoupled}}
  \end{subfigure}
  \caption{Feature decoupling using the PCA. The PCA identifies the directions with the highest data variance and enables the extraction of linearly independent features.}
  \label{fig:method-feature-pca}
\end{figure}

The PCA \cite{pearson1901liiilinesplanes} constitutes an unsupervised learning algorithm that decorrelated data, e.g., from \cref{fig:method-feature-pca-rotated}) to \cref{fig:method-feature-pca-decoupled}) \cite{goodfellow2016deeplearning}.
As a linear transformation, the PCA identifies an orthonormal basis by maximizing the variance in the directions of the principal components.
In the field of model order reduction, the PCA is often used to compute proper orthogonal decompositions, which eliminate undesirable frequencies from mechanical problems and reduce the model dimensionality to achieve computational speed-up (cf.\ \citeauthor{freitag2017recurrentneuralnetworks} \cite{freitag2017recurrentneuralnetworks}, \citeauthor{cao2016hybridrnngpodsurrogate} \cite{cao2016hybridrnngpodsurrogate}, or \cite{bamer2017efficientmontecarlo} \citeauthor{bamer2017efficientmontecarlo}).
Used in this explainable AI approach, the PCA is not used to accelerate computations, but to analyze and explain neural network behavior.

To compute the PCA on a dataset with correlated features (\cref{fig:method-feature-pca-rotated}), the dataset of $M$ samples $\ARR{x}_m$ is collected in a dataset tensor \smash{$\ARR{X}$} and centered feature-wise to \smash{$\widebar{\ARR{X}}$}:
\begin{align}
  \ARR{X} &= \brk[s]{\ARR{x}_1 \dots \ARR{x}_M}^{\mathsf{T}} \;,\\
  \label{eq:method-pca1}
  \widebar{\ARR{X}} &= \ARR{X} - \STAT{E}\brk[s]{\ARR{X}}
\end{align}
Applied to the centered dataset tensor $\widebar{\ARR{X}}$, the Singular Value Decomposition (SVD), 
\begin{equation}
  \label{eq:method-pca2}
  \operatorname{SVD}: \quad \widebar{\ARR{X}} \longmapsto \ARR{U}, \ARR{\Sigma}, \ARR{W}^{\mathsf{T}} \;,
\end{equation}
decomposes \smash{$\widebar{\ARR{X}}$} into the left-singular vectors \smash{$\ARR{U}$}, the singular-value diagonal matrix \smash{$\ARR{\Sigma}$}, and the right-singular vectors \smash{$\ARR{W}$}.
Conventionally, most SVD algorithms sort the singular vectors and singular values on the basis of the magnitude of the latter, where the highest singular value in \smash{$\ARR{\Sigma}$} corresponds to the highest variance (i.e., frequency) in the dataset.
Feature-wise contraction of the centered dataset $\widebar{\ARR{X}}$ with $\ARR{W}$ yields the decoupled data tensor \smash{$\bar{\ARR{x}}$}:
\begin{equation}
  \label{eq:method-pca3}
  \bar{\ARR{x}} = \widebar{\ARR{X}} \ARR{W}\;.
\end{equation}
The reciprocal square root of the singular values $\sqrt{\ARR{\Sigma}^{-1}}$ scales the decoupled dataset \smash{$\bar{\ARR{x}}$} down to unit variance.

To use the PCA on LSTM cell states $\ARR{c}_t$ from \cref{eq:nn-lstm-cell-flow}, the data tensor $\ARR{C}$ of sequence length $T$ is assembled:
\begin{equation}
  \label{eq:method-pca4}
  \ARR{C} \equiv \brk[s]1{\ARR{c}_1 \dots \ARR{c}_T} \;.
\end{equation}
\Cref{eq:method-pca1,eq:method-pca2,eq:method-pca3,eq:method-pca4} yield
\begin{equation}
  \widebar{\ARR{C}} = \bar{\ARR{C}} \ARR{W} \quad \text{with} \quad \widebar{\ARR{C}} = \ARR{C} - \STAT{E}\brk[s]{\ARR{C}} \;.
\end{equation}

The columns $\widebar{\ARR{C}}_t$ ($t = 1 \dots T$) of the decoupled state tensor $\widebar{\ARR{C}}$ describe the temporal behavior of the cell in the principal axes.
The associated singular value $\Sigma_{f}$ divided by the sum of all singular values quantifies the relative importance of the corresponding principal components, i.e., how much each principal component explains the cell state variance.
Often, most of the variance can be explained using the first three principal components, \smash{$\bar{\ARR{c}}^{\mathrm{I}}$}, \smash{$\bar{\ARR{c}}^{\mathrm{II}}$}, and \smash{$\bar{\ARR{c}}^{\mathrm{III}}$}, which describe the memory response of the majority of the cell units.
To investigate the ability to explain the mechanical problem, the neural network's major memory cell responses, \smash{$\bar{\ARR{c}}^{\mathrm{I}}$}, \smash{$\bar{\ARR{c}}^{\mathrm{II}}$}, and \smash{$\bar{\ARR{c}}^{\mathrm{III}}$}, are compared to the algorithmic history variables $\ARR{q}$.

This comparison demonstrates that the neural network understood temporal mechanical problems in line with the physically observed evolution laws.

Note that the ability to generalize, i.e., to achieve a reproducible and equally accurate result on unknown data, as outlined in \cref{ss:methods-hyperparameter-search}, remains independent of the ability to model the result on correct physical assumptions, as described in this subsection.
To achieve generalizable results, systematic hyperparameter tuning thus is the necessary prerequisite for the explainable AI approaches.

\section{Fundamental case studies}
\label{sec:results}

The following three case studies demonstrate the systematic hyperparameter search strategy and the new explainable AI approach.
First, the proposed systematic hyperparameter search strategy will be demonstrated in the scope of a case study for an intelligent nonlinear elastic constitutive model.
Thereafter, the latter two case studies combine the systematic hyperparameter search strategy with explainable AI, in order to interpret inelastic time-variant constitutive behavior.

For all data-driven constitutive models, the same data-generation strategy provides training, validation, and test data. 
For a sequence length of $T = \num{10000}$ and \smash{$\Omega = 5$} phases of loading and unloading, we sample control values from a random normal distribution \smash{$\DIST{N}(\mu=0, \sigma=1)$}.
Between those control values, a variety of ramping functions, e.g., linear, quadratic, or sinusoidal, interpolate the intermediate values to assemble stress- or strain-controlled loading sequences.
The ramping functions are selected to cover a variety of constitutive responses based on the investigated constitutive models.
Thereafter, the numerical implementation of each reference constitutive model generates \smash{$M^{\text{total}} = \num{10000}$} samples, which are split randomly $\SI{70}{\percent}-\SI{15}{\percent}-\SI{15}{\percent}$ into training, validation, and test set.

Three constitutive models are selected to demonstrate fundamental mechanical material behavior, including finite deformations, long-term temporal behavior, and rate dependency.
To enhance interpretability, each one-dimensional constitutive model uses dimensionless and purely academic parameter values.
Using preprocessing strategies, such as normalization and augmenting the input with explicit model parameters, models with arbitrary material parameter values can be created \cite{koeppe2018efficientnumericalmodeling,koeppe2020intelligentnonlinearmeta}.
First, an incompressible Neo-Hooke constitutive model ($E = \num{1.0}$) computes the nonlinear stress response to stretch-controlled loading.
Due to the hyperelastic problem, all loading and unloading phases $\omega = 1 \dots \Omega$ use linear ramps.
The investigated time-distributed dense neural network architectures receive the stretch $\uplambda$ as input and learn to predict the Cauchy stress $\upsigma$.
Second, a perfect Prandtl-Reuss elastoplasticity model ($E = 1$, $\upsigma_Y = 0.6$) is subjected to strain-controlled loading.
The phase interpolation functions are sampled randomly from linear, quadratic, square-root, exponential, sine, and half-sine ramping functions.
The investigated RNN architectures use the strain \smash{$\upvarepsilon$} to return the stress \smash{$\upsigma$} and plastic strain $\upvarepsilon^{\mathrm{P}}$.
Finally, a Poynting-Thomson constitutive model, defined by \smash{$E = \num{1.0}$}, \smash{$E_1 = \num{0.5}$}, and \smash{$\tau_1 = \num{0.1667}$} is integrated in time ($\mathrm{T} = \num{1}$) with an implicit backward Euler-scheme.
The stress-controlled phases $\omega = 1 \dots \Omega$ include linear ramps and constant phases to investigate creeping behavior.
The RNN architectures process the stress \smash{$\upsigma$} and return the strain $\upvarepsilon$ and the viscous branch stress \smash{$\upsigma_1$}.
For all neural networks, a final time-distributed dense layer of output size applies a linear transformation to cover the entire range of real values.

For the neural network architecture and the hyperparameters, the Hyperband algorithm \cite{li2017hyperbandnovelbanditbased} systematically explores and investigates the high-dimensional search domain.
During the training of each configuration, the Adam algorithm \cite{kingma2014adammethodstochastic} minimizes the MSE on the training set ($N = 51$ epochs) and reports the validation loss to evaluate the configuration in the scope of Hyperband.
To avoid artificially penalizing specific weight values, neither L\textsuperscript{1} nor L\textsuperscript{2} regularization is used, optimizing the unconstrained MSE.
In the Hyperband brackets, each step performed by the Successive Halving algorithm eliminates the worst configurations from all configurations by a factor of $\eta = \num{3.7}$.
In \Cref{tab:results-constitutive-hyperparameter-search-domain}, the search domains for the recurrent and dense neural network architectures are described. 
After the search, we train the best configuration for the full duration of $N = 301$ epochs.
The resulting parameter values $\ARR{\theta}^{\text{opt}}$ are used to evaluate the test set to compute the test errors.

\begin{table}[htbp]
  \caption{Hyperband search domain for constitutive models.}
  \label{tab:results-constitutive-hyperparameter-search-domain}
  \centering
  \begin{tabulary}{\linewidth}{LC>{\centering\arraybackslash}p{0.67\textwidth}}
    \toprule
    \textbf{Variable} & & \textbf{Search domain}\\
    \midrule
    layer width & $d^{(l)}$ & $\brk[s]*{4, 8, \dots, 128}$ \\
    neural network depth & $L$ & $\brk[s]*{2, 3, 4, 5, 6}$ \\
    base learning rate & $\alpha$ & $\brk[s]*{\num{1e-2}, \num{3e-3}, \num{1e-3}, \num{3e-4}, \num{1e-4}}$\\
    batch size & $M$ & $\brk[s]*{32, 64, 128}$\\
    \midrule
    a) activation {\footnotesize (dense only)} & $\FUN{f}$ & $\brk[s]*{\mathrm{rect}, \mathrm{sig}, \mathrm{tanh}, \mathrm{elu}, \mathrm{splus}}$ \\
    b) cell type {\footnotesize (recurrent only)} &  & $\brk[s]*{\mathrm{LSTM}, \mathrm{GRU}, \mathrm{recurrent-tanh}, \mathrm{recurrent-rect}}$\\
    \bottomrule
  \end{tabulary}
\end{table}

For the last two case studies, the novel explainable neural network approach (\Cref{ss:methods-explainable-ai}), employing the PCA on the cell states, analyzes the best RNN architectures to explain the intelligent inelastic constitutive models.

The Kadi4Mat \cite{brandt2021kadi4matresearchdata} data management infrastructure stores, links, and describes the data generated in this publication.
Published via the direct integration of Zenodo \cite{koeppe2021dataset}, the dataset includes the generated reference constitutive model dataset, the associated metadata of the constitutive and neural network models, and the serialized trained neural networks for different hyperparameter configurations at different epochs during training.
The data storage with the associated metadata and connections enhances the systematic hyperparameter search strategy with the option for long-term data sustainability, e.g., by reusing previous search results and hyperparameter configurations for future models.
Furthermore, the links between the dataset and the serialized neural network models provide a starting point for a bottom-up data ontology for machine learning in mechanics and material sciences.
Subsequently formalized, the data ontology will provide semantic rules that describe the relations and workflows inherent to the research data, which will enable additional analysis and explainable AI approaches.

\subsection{Systematic investigation of data-driven hyperelastic constitutive models}
\label{ss:results-hyperelasticity}

\begin{figure}[htbp]
  \centering
  \includetikz{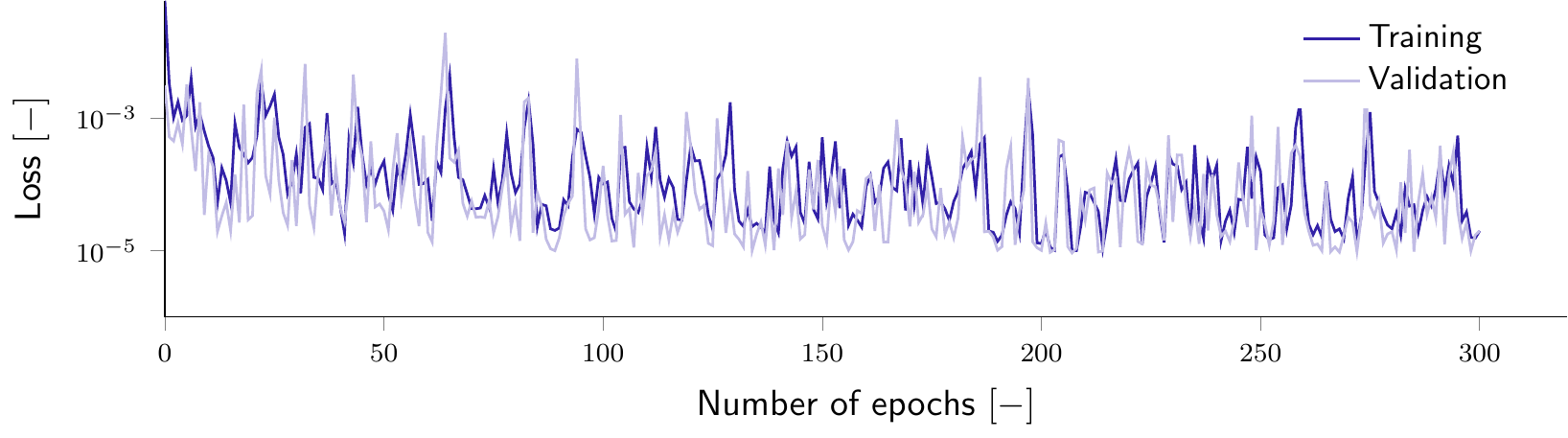}
  \vspace*{-0.5cm}
  \caption{Training and validation loss during hyperelastic constitutive model training. The strong oscillations can be attributed to the singular material behavior, at a stretch approaching zero.}
  \label{fig:results-constitutive-hyperelastic-loss-convergence}
\end{figure}

\begin{figure}[htbp]
  \centering
  \includetikz{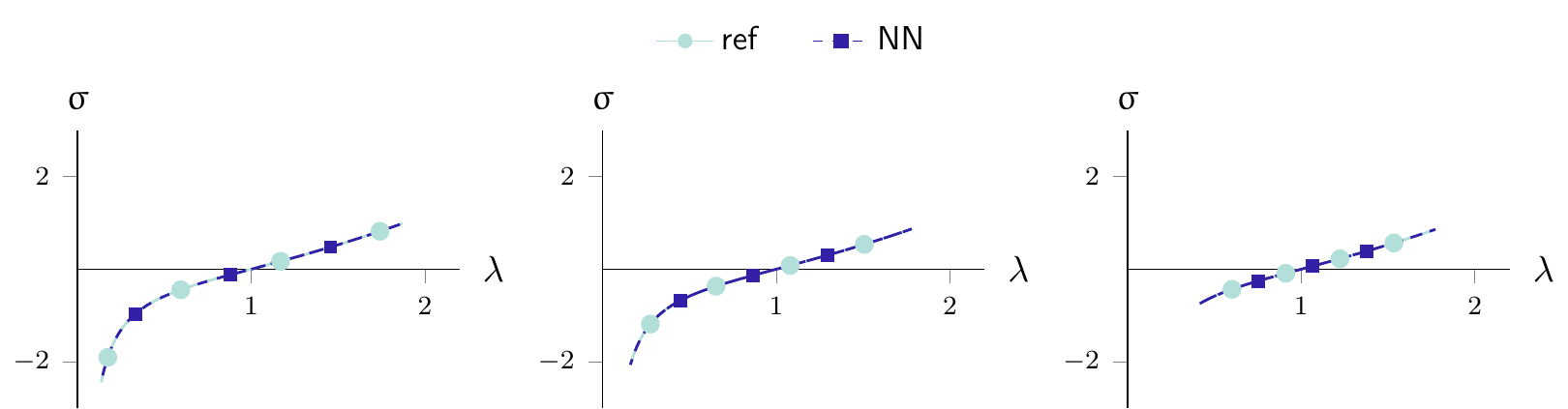}
  \caption{Three randomly selected stress-stretch curves for hyperelastic constitutive behavior. The data-driven constitutive model exactly matches the reference solution.}
  \label{fig:results-constitutive-hyperelasticity-stress-strain}
\end{figure}

For the hyperelastic problem, the best-performing hyperparameters were a batch size of $M = \num{64}$ and a base learning rate of $\alpha = \num{3e-3}$.
A deep feedforward neural network with $\num{5}$ hidden layers of width $\num{112}$ and rectifier activation achieved the best performance.
After training (\cref{fig:results-constitutive-hyperelastic-loss-convergence}) for $\num{301}$ epochs, the neural network achieves an MSE \smash{$\epsilon^{\text{MSE}}$} of $\num{1.275e-05}$ on the training, $\num{1.006e-5}$ on the validation, and $\num{1.015e-5}$ on the test set.
The same order of error magnitude on all three datasets indicates that the neural network achieved generalization.
\Cref{fig:results-constitutive-hyperelasticity-stress-strain} visualizes three randomly selected test samples.
For the one-dimensional hyperelastic case, the data-driven constitutive model is in perfect agreement with the reference model.

\subsection{Explaining data-driven elastoplastic constitutive models}
\label{ss:results-elastoplasticity}

For the elastoplastic problem, the systematic search strategy identified a best-performing architecture, characterized by a single LSTM cell, with a width of $\num{52}$ units followed by a linear time-distributed dense layer with $\num{2}$ units.
For the training, a batch size of $M = \num{32}$ and a base learning rate of $\alpha = \num{1e-3}$ resulted in the best result without overfitting, i.e., an MSE \smash{$\epsilon^{\text{MSE}}$} of $\num{5.724e-5}$ on the training, $\num{5.114e-5}$ on the validation, and $\num{4.375e-5}$ on the test set.
 
\begin{figure}[htbp]
  \centering
  \includetikz{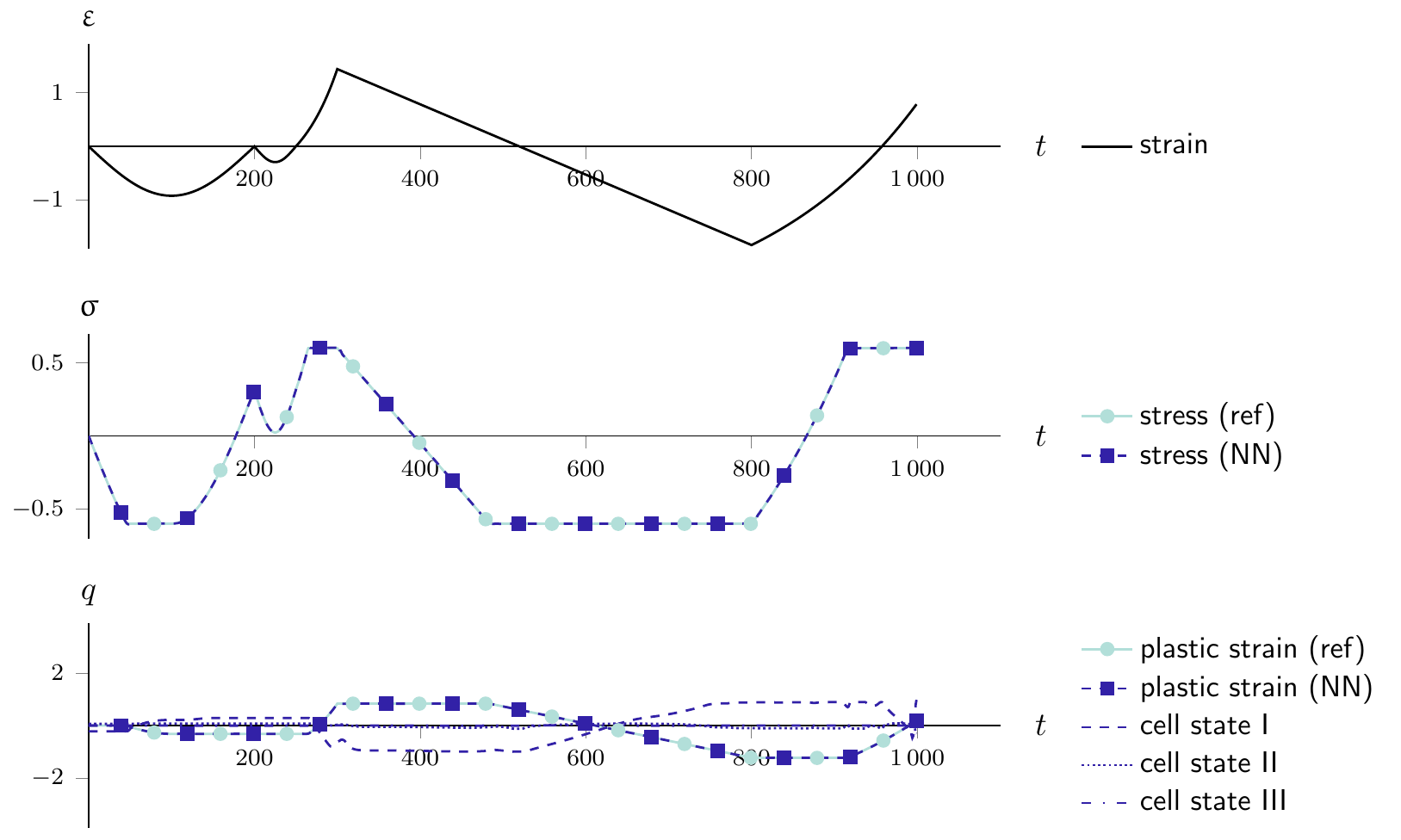}
  \caption{An explainable elastoplastic constitutive model. (Top) The strain-driven loading over the time increments. (Middle) The stress response of the reference and data-driven constitutive model. (Bottom) The plastic strain, compared to the three major principal components of the cell state. The cell states approximate the plastic strain (with a negative sign remedied by the output layer).}
  \label{fig:results-constitutive-elastoplasticity-explainable}
\end{figure}

To interpret and explain the RNN behavior, a random sample is extracted from the test set and evaluated using the data-driven constitutive model.
The explainable AI approach uses the PCA on the concatenated cell states of all four LSTMs and yields the three principal components with the highest singular values.
Expressed in the three major principal components and divided by their singular values, the cell states I, II, and III represent the joint response of the LSTM cell states (\Cref{sec:method-explaining-mechanical-neural-networks}).
\Cref{fig:results-constitutive-elastoplasticity-explainable} visualizes and compares the evolution of the stresses, strains, and history variables over the entire loading sequence.

The neural network's output predictions and the reference constitutive model match exactly, i.e., the stresses and plastic strain predictions are accurate.
By comparing the plastic strain with the internal cell states, the LSTM cell's decision-making behavior becomes apparent.
Without being trained directly, the cell states' principal components learn to approximate the plastic strain evolution (with a negative sign).
Since the PCA chooses the principal directions based on the variance, and the output layer can apply arbitrary weighting, the negative sign of the cell state does not affect the result.
The second and third principal components do not contribute to the joint cell state response.

\begin{figure}[htbp]
  \centering
  \includetikz{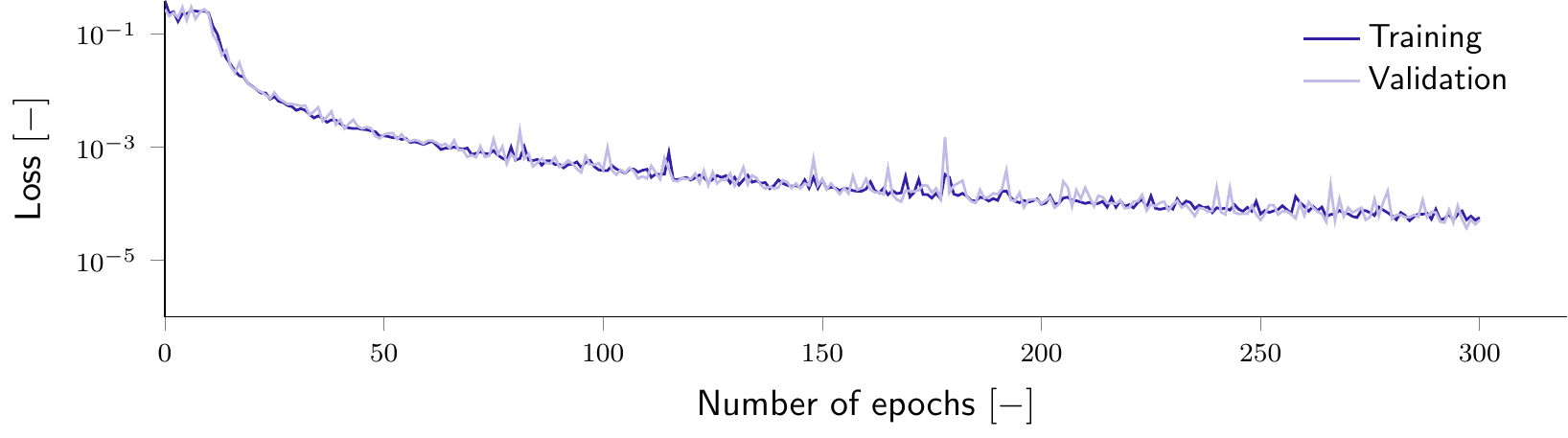}
  \vspace*{-0.5cm}
  \caption{Training and validation loss during elastoplastic constitutive model training.
  }
  \label{fig:results-constitutive-elastoplastic-loss-convergence}
\end{figure}
 
\subsection{Explaining data-driven viscoelastic constitutive models}
\label{ss:results-viscoelasticity}

\begin{figure}[htbp]
  \centering
  \includetikz{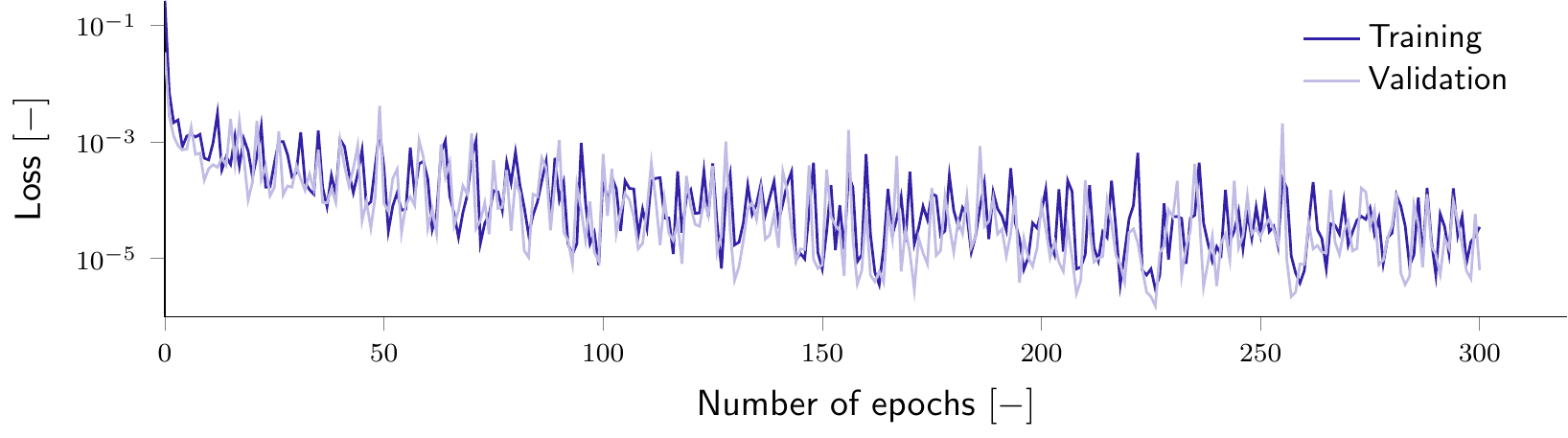}
  \vspace*{-0.5cm}
  \caption{Training and validation loss during viscoelastic constitutive model training.
  }
  \label{fig:results-constitutive-viscoelastic-loss-convergence}
\end{figure}

For the viscoelastic problem, the Hyperband search found the best-performing architecture to be three GRU cells with $\num{120}$ units, followed by a time-distributed dense layer ($\num{2}$ units and linear activation).
For training convergence, a batch size of $M = \num{32}$ and a base learning rate of $\alpha = \num{1e-3}$ achieved the best results.
After training for the full $\num{301}$ epochs (\cref{fig:results-constitutive-viscoelastic-loss-convergence}), the neural network achieved a best MSE \smash{$\epsilon^{\text{MSE}}$} of $\num{2.053e-7}$ on the training, $\num{7.364e-7}$ on the validation, and $\num{1.149e-6}$ on the test set, indicating generalization and an approximation near machine precision ($\sim\!\order{10^7}$ for single-precision floating-point arithmetic).

\begin{figure}[htbp]
  \centering
  \includetikz{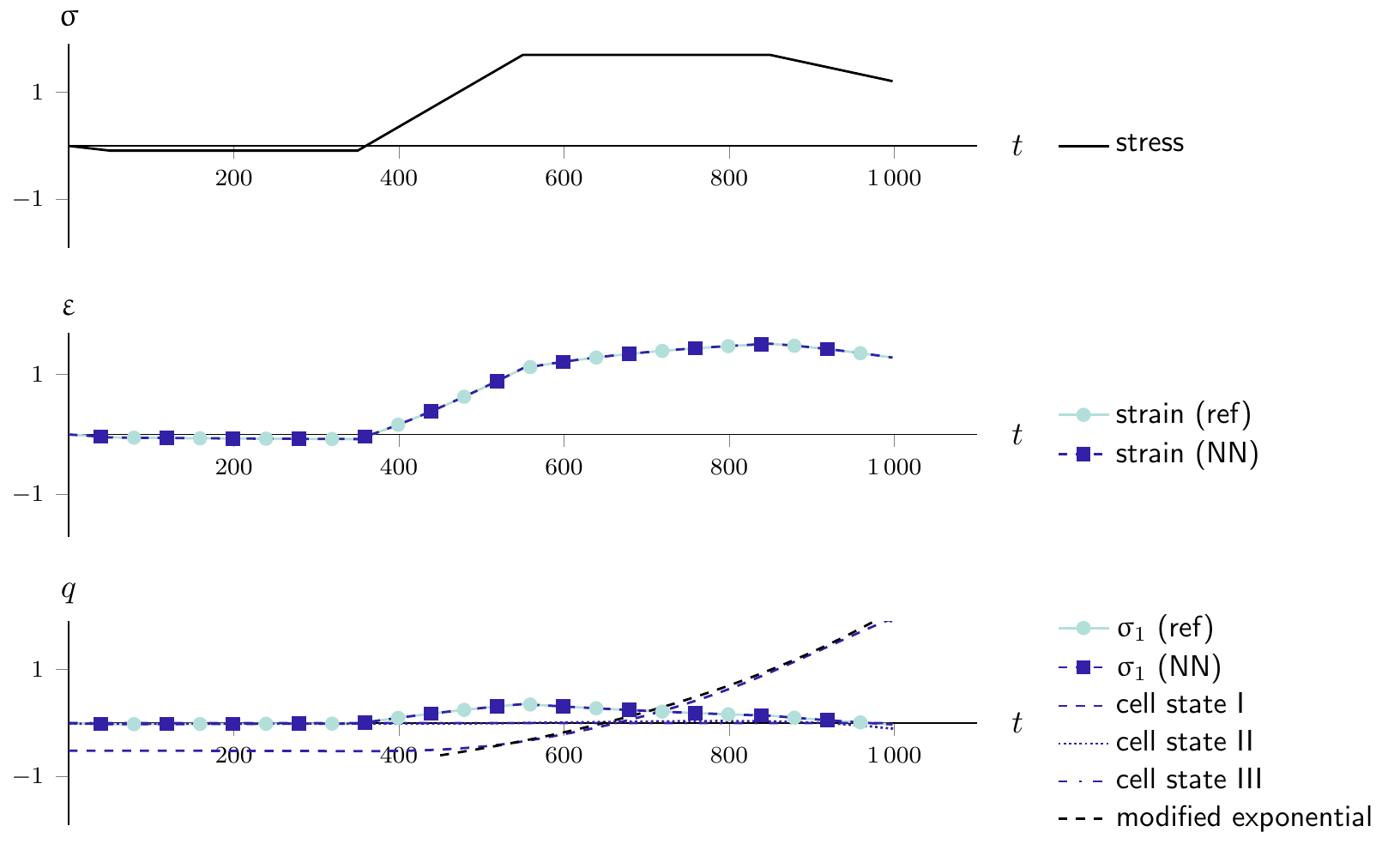}
  \caption{An explainable viscoelastic constitutive model. (Top) The stress-driven loading over time. (Middle) The strain response of the reference and data-driven constitutive model. (Bottom) The branch history stress compared to the three major principal components of the cell state. Instead of mimicking the history variables used to generate training data, the cell states learned a generic solution for viscoelasticity: a modified exponential function that can be shifted and scaled at will by the output layer.}
  \label{fig:results-constitutive-viscoelastic-explainable}
\end{figure}

\Cref{fig:results-constitutive-viscoelastic-explainable} exemplifies the explainable AI strategy on one randomly generated sample that was not used for training or validation.
In the top row, the stress is plotted over the increments, which constitutes the input to the reference constitutive and neural network models.
The second row compares the strain response of the reference and neural network models.
The final row depicts the history variables $q$. 
The branch stress \smash{$\upsigma_1$} is computed by both the reference and data-driven constitutive models.
Both strain and branch stress match accurately, as indicated by the low test loss.
Finally, the PCA on the GRU cell state over time yields the data-driven constitutive history variables, i.e., the three principal components with the highest singular values, which govern the major evolution of the data-driven constitutive model.

Instead of approximating the history variables \smash{$\upsigma_1$}, used to generate the training data with backward Euler time integration, the first principal component of the cell states approximates a modified exponential function, correctly identifying the exponential behavior of viscoelasticity. 
Shifting, scaling, and changing the sign of a function represent trivial operations for previous and subsequent neural network layers.
Thus, the weights of the RNN and the last dense layer can modify the cell output at will to assemble a solution, such as \cref{eq:method-viscoelastic-creep-closed-form}.
The second and third principal components do not contribute to the joint cell state response.
 
\subsection{Discussion}
\label{ss:results-viscoelasticity}

The objectives of this publication were to systematically search neural network architectures for data-driven constitutive models of fundamental material behavior and to explain the trained neural networks.

The test errors (\num{2.446e-6} to \num{4.375e-5}) approach machine precision for the single-precision floating-point arithmetic used to compute them.
The neural network performance surpasses the elastoplastic constitutive model for uniaxial tension and compression in the recent works of \citeauthor{huang2020machinelearningbased} \cite{huang2020machinelearningbased} (\num{3.93e-2}) and \citeauthor{alwattar2019developmentelasticmaterial} \cite{alwattar2019developmentelasticmaterial} ($\sim\!\num{1e-5}$, as reported on the training set).

Even beyond mechanics, applying the PCA to explain neural network cell states constitutes a novel and promising explainable AI method.
The PCA investigation proposed in this work can explain the recurrent cell state behavior, despite the considerable size of the investigated architecture, compared to the one-dimensional fundamental problems.
Due to the quality of the results and the explainability of the neural network, `Clever-Hans' predictors \cite{lapuschkin2019unmaskingcleverhans} can be ruled out.

As the first explainable neural network approach in mechanics, the case studies demonstrate how neural networks can help to explain material behavior.
For the elastoplastic problem, the recurrent cell state identified the same history variables used in computing the ground truth.
For the viscoelastic problem, the data generation used numerical time-integration, but the neural network found a solution similar to the exponential closed-form solution.
Therefore, if the ground truth is not known, e.g., when learning from raw experimental data, explainable neural network approaches can possibly identify underlying closed-form solutions.
By characterizing new materials or material behavior, e.g., at extreme loading conditions, explainable AI can help guide researchers in mechanics and materials sciences towards new analytic closed-form solutions that elegantly model the materials in the desired ranges.

The results were achieved inductively, with minimal user input, as the first mechanical application of a Hyperband-inspired systematic search strategy.
To further improve the results of the neural networks, several approaches are conceivable, most of which are deductively derived from deep learning and mechanical domain knowledge. 
Ensemble learning and related regularizers, such as dropout \cite{srivastava2014dropoutsimpleway}, for example, are known to improve the quality of the results further \cite{goodfellow2016deeplearning,srivastava2014dropoutsimpleway}.
Similarly, physics-informed and physics-guided approaches, as proposed by \citeauthor{raissi2019physicsinformedneuralnetworks} \cite{raissi2019physicsinformedneuralnetworks}, \citeauthor{yang2019adversarialuncertaintyquantification} \cite{yang2019adversarialuncertaintyquantification}, or
\citeauthor{kissas2020machinelearningcardiovascular} \cite{kissas2020machinelearningcardiovascular}, use mechanical domain knowledge to improve neural network models of physical systems.
In contrast, the data-driven constitutive models trained in this work represent an inductive, fundamental, and accurate approach that offers intuition for possible neural network architectures and hyperparameter configurations for higher-dimensional mechanical problems.
In combination with the explainable AI approach, the data-driven constitutive models represent the first step towards \emph{physics-informing} neural networks.

\section{Concluding remarks and outlook}
\label{sec:conclusion}

We proposed a step towards \emph{physics-informing} neural networks, which inductively complement existing deductive approaches for physics-informed and physics-guided neural networks.
To that end, a systematic hyperparameter search strategy was implemented to identify the best neural network architectures and training parameters efficiently.
For the analysis of the best neural networks, we proposed a novel explainable AI approach, which uses a PCA to explain the distributed representations in the cell states of RNNs.

The search strategy and explainable AI approach were demonstrated on data-driven constitutive models that learned fundamental material behavior, i.e., one-dimensional hyperelasticity, elastoplasticity, and viscoelasticity.
For all case studies, the best neural network architectures achieved test errors in the order of $\sim\!\numrange{1e-5}{1e-6}$.
In particular, for hyperelasticity, the test error approached machine precision, despite the singular behavior for stretches approaching zero.
For elastoplasticity, the novel explainable AI approach identified that the recurrent cell states learned history variables equivalent to the plastic strain, i.e., the history variables used to generate the original data.
Remarkably, for viscoelasticity, the explainable AI approach found that the best performing neural network architecture used an exponential function as the basis for its decisions instead of the algorithmic history variables used to generate the training data.

These findings imply that systematic hyperparameter search, coupled with explainable AI, can help identify and characterize numerical and analytical closed-form solutions for constitutive models independent of the data origin.
Thus, new materials can potentially be characterized with data originating from experiments, using the approach proposed in this work.

Future studies will apply and extend the proposed strategies to more complex material models.
Of particular interest are viscoelastic materials subjected to strain rates that cover multiple decades, where conventional numerical models require numerous algorithmic history variables.
Eventually, new materials, where the analytic closed-form solutions are as-of-yet unknown and numerical solutions are challenging to implement, can be characterized with developments based on the present work.
Finally, applications beyond constitutive models are conceivable.
For spatio-temporal problems, e.g., \cite{koeppe2020intelligentnonlinearmeta}, the explainable AI approach outlined in this work needs to be extended to leverage the spatial structure.

In forthcoming work, it is intended to extend the developed artificial intelligence approach within the Kadi4Mat \cite{brandt2021kadi4matresearchdata} framework to higher dimensional data containing 2D and 3D spatial plus temporal information to predict microstructure-mechanics correlations. 
The database used to train the neural network algorithms relies on digital twin data from synchronously conducted experiments and simulations of mechanically loaded polycrystalline and multiphase materials. 
Based on the training, the AI approach is applied to large-scale micromechanics-microstructure simulations so as to provide new insights, e.g., into mechanically induced nucleation events of new phases and grain variants or into microcrack probabilities. 
The combination of new AI concepts and advanced high-performance materials simulations shall establish an integral component of the research data infrastructure to enable computational methods for an accelerated design of new materials. 

\section*{Acknowledgements}

The authors gratefully acknowledge financial support by the Federal Ministry of Education and Research (BMBF) in the project FestBatt (project number 03XP0174E) and by the Ministry of Science, Research and Art Baden-Württemberg in the project MoMaF - Science Data Center, with funds from the state digitization strategy digital@bw (project number 57).
The authors thank Leon Geisen for his editorial support.

\bibliographystyle{elsarticle-num-names}
 \bibliography{references}





\end{document}

%% file: packages.tex
\usepackage[T1]{fontenc}
\usepackage[utf8]{inputenc}

\usepackage{xcolor}
  \definecolor{vaiolet}{RGB}{50,33,167}
  \definecolor{kitblack}{RGB}{0,0,0}
  \definecolor{kitwhite}{RGB}{255,255,255}
  \definecolor{kitgray}{RGB}{64,64,64}
  \definecolor{kitgreen}{RGB}{0,150,130}
  \definecolor{kitblue}{RGB}{70,100,170}
  \definecolor{kitmaygreen}{RGB}{140,182,60}
  \definecolor{kityellow}{RGB}{252,229,0}
  \definecolor{kitorange}{RGB}{223,155,27}
  \definecolor{kitbrown}{RGB}{167,130,46}
  \definecolor{kitred}{RGB}{162,34,35}
  \definecolor{kitmagenta}{RGB}{163,16,124}
  \definecolor{kitcyan}{RGB}{35,161,224}
  \definecolor{rwthblack}{RGB}{0,0,0}
  \definecolor{rwthwhite}{RGB}{255,255,255}
  \definecolor{rwthgray}{RGB}{51,51,51}
  \definecolor{rwthlightgray}{RGB}{204,204,204}
  \definecolor{rwthblue}{RGB}{0,84,159}
  \definecolor{rwthlightblue}{RGB}{142,186,226}
  \definecolor{rwthsuperlightgray}{RGB}{247,247,247}
  \definecolor{rwthpetrol}{RGB}{0,97,101}
  \definecolor{rwthteal}{RGB}{0,152,161}
  \definecolor{rwthmaygreen}{RGB}{189,205,0}
  \definecolor{rwthgreen}{RGB}{87,171,39}
  \definecolor{rwthyellow}{RGB}{255,237,0}
  \definecolor{rwthorange}{RGB}{246,168,0}
  \definecolor{rwthmagenta}{RGB}{227,0,102}
  \definecolor{rwthred}{RGB}{204,7,30}
  \definecolor{rwthlightred}{RGB}{230,150,121}
  \definecolor{rwthbordeaux}{RGB}{161,16,53}
  \definecolor{rwthviolet}{RGB}{97,33,88}
  \definecolor{rwthpurple}{RGB}{122,111,172}
  \definecolor{mplc0}{RGB}{31,119,180}
  \definecolor{mplc1}{RGB}{255,127,14}
  \colorlet{pairedcol1a}{rwthblue}
  \colorlet{pairedcol1b}{rwthlightblue}
  \colorlet{pairedcol2a}{rwthorange}
  \colorlet{pairedcol2b}{rwthorange!50!rwthwhite}
  \colorlet{pairedcol3a}{rwthgreen}
  \colorlet{pairedcol3b}{rwthgreen!50!rwthwhite}
  \colorlet{pairedcol4a}{rwthbordeaux}
  \colorlet{pairedcol4b}{rwthbordeaux!50!rwthwhite}
  \colorlet{plotcol1}{rwthblue}
  \colorlet{plotcol2}{rwthorange}
  \colorlet{plotcol3}{rwthgreen}
  \colorlet{plotcol4}{rwthbordeaux}
  \colorlet{plotcol5}{rwthlightblue}
  \colorlet{plotcol6}{rwthorange!50!rwthwhite}
  \colorlet{plotcol7}{rwthgreen!50!rwthwhite}
  \colorlet{plotcol8}{rwthbordeaux!50!rwthwhite}

\usepackage{amsmath}
\usepackage{amsfonts}
\usepackage{amssymb}
\usepackage{mathtools}
    \DeclarePairedDelimiter\ceil{\lceil}{\rceil}
    \DeclarePairedDelimiter\floor{\lfloor}{\rfloor}
\usepackage{upgreek}    
\usepackage{accents}    
    \DeclareFontFamily{U}{mathx}{\hyphenchar\font45}
    \DeclareFontShape{U}{mathx}{m}{n}{<-> mathx10}{}
    \DeclareSymbolFont{mathx}{U}{mathx}{m}{n}
    \DeclareMathAccent{\widebar}{0}{mathx}{"73}
\usepackage{siunitx}    
\usepackage[mathscr]{eucal} 
\usepackage{bbm}            
\usepackage{bm}
    \newcommand{\ARR}[1]{\bm{#1}}
    
    \newcommand{\FUN}[1]{\mathrm{#1}}
    \newcommand{\FNCTL}[1]{\mathcal{#1}}
    
    \newcommand{\STAT}[1]{\mathbb{#1}}
    \newcommand{\PROB}[1]{\mathrm{#1}}
    \newcommand{\DIST}[1]{\mathcal{#1}}
\usepackage{relsize}        
\usepackage{exscale}        
\usepackage{nicefrac}       
\usepackage{physics}
\usepackage{delimset}       

\usepackage{graphicx}
\usepackage{tikz}
    \usetikzlibrary{decorations}
    \usetikzlibrary{decorations.markings}
    \usetikzlibrary{decorations.text}
    \usetikzlibrary{shapes.geometric}
    \usetikzlibrary{shapes}
    \usetikzlibrary{arrows}
    \usetikzlibrary{arrows.meta}
    \usetikzlibrary{chains}
    \usetikzlibrary{matrix}
    \usetikzlibrary{calc}
    \usetikzlibrary{fit}
    \usetikzlibrary{backgrounds}
    \usetikzlibrary{3d}
    \usetikzlibrary{shadows}
    \usetikzlibrary{positioning}
    \usetikzlibrary{bending}
    \usetikzlibrary{intersections}
    \usetikzlibrary{external}
    \newcommand{\includetikz}[2][]{%
      \includegraphics[#1]{tikz-compiled/#2.pdf}%
    }
\usepackage{pgffor}
\usepackage{pgfplots}
    \usepgfplotslibrary{groupplots}
    \usepgfplotslibrary{fillbetween}
    \pgfplotsset{
      compat=1.15,
      my centered axis style/.style={
        axis x line=center,
        axis y line=center,
        axis line style={-},
        xlabel style={
            at={(ticklabel* cs:1.025)},
            anchor=west,
            font=\small\sffamily,
        },
        ylabel style={
            at={(ticklabel* cs:1.025)},
            anchor=south,
            font=\small\sffamily,
            rotate=0,
        },
        tick align=outside,
        scaled y ticks = false,
        y tick label style={
          font=\scriptsize, 
          /pgf/number format/.cd,%
            set thousands separator={\,},
            fixed,
        },
        scaled x ticks = false,
        x tick label style={
          font=\scriptsize,
          /pgf/number format/.cd,%
            set thousands separator={\,},
            fixed,
        },
        max space between ticks=50pt,
        try min ticks=3,
        legend cell align=left,
        legend style={draw=none, nodes={font=\smaller\sffamily}},
      },
      my botleft axis style/.style={
        axis x line=bottom,
        axis y line=left,
        axis line style={-},
        xlabel near ticks,
        xlabel style={
            font=\small\sffamily,
        },
        ylabel near ticks,
        ylabel style={
            font=\small\sffamily,
        },
        tick align=outside,
        scaled y ticks = false,
        y tick label style={
          font=\scriptsize, 
          /pgf/number format/.cd,%
            set thousands separator={\,},
            fixed,
        },
        scaled x ticks = false,
        x tick label style={
          font=\scriptsize,
          /pgf/number format/.cd,%
            set thousands separator={\,},
            fixed,
        },
        max space between ticks=50pt,
        try min ticks=3,
        legend cell align=left,
        legend style={draw=none, nodes={font=\smaller\sffamily}},
      },
    }

\usepackage{booktabs}
\usepackage{longtable}
\usepackage{tabulary}
\usepackage{multicol}
\usepackage{multirow}

\usepackage[ruled, vlined]{algorithm2e}
    \newcommand{\alignedintertext}[1]{%
      \noalign{%
        \vtop{\hsize=\linewidth#1\par\vspace*{-3.5ex}
        \expandafter}%
        \expandafter\prevdepth\the\prevdepth
      }%
    }

\usepackage{float}
\usepackage[labelfont=bf,justification=raggedright,singlelinecheck=false]{caption}
\usepackage{subcaption}

\usepackage[colorlinks, allcolors=kitblue]{hyperref}
\usepackage[capitalize]{cleveref}
 \crefname{subsection}{Subsection}{Subsections}
 \Crefname{subsection}{Subsection}{Subsections}
 \crefname{equation}{}{}
 \Crefname{equation}{}{}
 \crefname{figure}{Fig.}{Figs.}
 \Crefname{figure}{Fig.}{Figs.}
 \crefname{tabular}{Tab.}{Tabs.}
 \Crefname{tabular}{Tab.}{Tabs.}



%% file: fig/constitutive-rheological-model-hyperelasticity.pdf_tex
\begingroup%
  \makeatletter%
  \providecommand\color[2][]{%
    \errmessage{(Inkscape) Color is used for the text in Inkscape, but the package 'color.sty' is not loaded}%
    \renewcommand\color[2][]{}%
  }%
  \providecommand\transparent[1]{%
    \errmessage{(Inkscape) Transparency is used (non-zero) for the text in Inkscape, but the package 'transparent.sty' is not loaded}%
    \renewcommand\transparent[1]{}%
  }%
  \providecommand\rotatebox[2]{#2}%
  \newcommand*\fsize{\dimexpr\f@size pt\relax}%
  \newcommand*\lineheight[1]{\fontsize{\fsize}{#1\fsize}\selectfont}%
  \ifx\svgwidth\undefined%
    \setlength{\unitlength}{226.77165354bp}%
    \ifx\svgscale\undefined%
      \relax%
    \else%
      \setlength{\unitlength}{\unitlength * \real{\svgscale}}%
    \fi%
  \else%
    \setlength{\unitlength}{\svgwidth}%
  \fi%
  \global\let\svgwidth\undefined%
  \global\let\svgscale\undefined%
  \makeatother%
  \begin{picture}(1,0.75)%
    \lineheight{1}%
    \setlength\tabcolsep{0pt}%
    \put(0,0){\includegraphics[width=\unitlength,page=1]{constitutive-rheological-model-hyperelasticity.pdf}}%
    \put(0.05622804,0.47302632){\makebox(0,0)[lt]{\lineheight{1.25}\smash{\begin{tabular}[t]{l}$\upsigma$\end{tabular}}}}%
    \put(0.86435671,0.47302632){\makebox(0,0)[lt]{\lineheight{1.25}\smash{\begin{tabular}[t]{l}$\upsigma$\end{tabular}}}}%
    \put(0.49610667,0.52905524){\makebox(0,0)[t]{\lineheight{1.25}\smash{\begin{tabular}[t]{c}$\mu$\end{tabular}}}}%
  \end{picture}%
\endgroup%

%% file: fig/constitutive-rheological-model-viscoelasticity.pdf_tex
\begingroup%
  \makeatletter%
  \providecommand\color[2][]{%
    \errmessage{(Inkscape) Color is used for the text in Inkscape, but the package 'color.sty' is not loaded}%
    \renewcommand\color[2][]{}%
  }%
  \providecommand\transparent[1]{%
    \errmessage{(Inkscape) Transparency is used (non-zero) for the text in Inkscape, but the package 'transparent.sty' is not loaded}%
    \renewcommand\transparent[1]{}%
  }%
  \providecommand\rotatebox[2]{#2}%
  \newcommand*\fsize{\dimexpr\f@size pt\relax}%
  \newcommand*\lineheight[1]{\fontsize{\fsize}{#1\fsize}\selectfont}%
  \ifx\svgwidth\undefined%
    \setlength{\unitlength}{226.77165354bp}%
    \ifx\svgscale\undefined%
      \relax%
    \else%
      \setlength{\unitlength}{\unitlength * \real{\svgscale}}%
    \fi%
  \else%
    \setlength{\unitlength}{\svgwidth}%
  \fi%
  \global\let\svgwidth\undefined%
  \global\let\svgscale\undefined%
  \makeatother%
  \begin{picture}(1,0.5)%
    \lineheight{1}%
    \setlength\tabcolsep{0pt}%
    \put(0,0){\includegraphics[width=\unitlength,page=1]{constitutive-rheological-model-viscoelasticity.pdf}}%
    \put(0.0590792,0.30350038){\makebox(0,0)[lt]{\lineheight{1.25}\smash{\begin{tabular}[t]{l}$\upsigma$\end{tabular}}}}%
    \put(0.85282917,0.30350038){\makebox(0,0)[lt]{\lineheight{1.25}\smash{\begin{tabular}[t]{l}$\upsigma$\end{tabular}}}}%
    \put(0,0){\includegraphics[width=\unitlength,page=2]{constitutive-rheological-model-viscoelasticity.pdf}}%
    \put(0.63873363,0.19242189){\makebox(0,0)[t]{\lineheight{1.25}\smash{\begin{tabular}[t]{c}$\tau_1$\end{tabular}}}}%
    \put(0.38483372,0.19242189){\makebox(0,0)[t]{\lineheight{1.25}\smash{\begin{tabular}[t]{c}$E_1$\end{tabular}}}}%
    \put(0.49685826,0.45700521){\makebox(0,0)[t]{\lineheight{1.25}\smash{\begin{tabular}[t]{c}$E$\end{tabular}}}}%
  \end{picture}%
\endgroup%

%% file: fig/constitutive-rheological-model-elastoplasticity-ip.pdf_tex
\begingroup%
  \makeatletter%
  \providecommand\color[2][]{%
    \errmessage{(Inkscape) Color is used for the text in Inkscape, but the package 'color.sty' is not loaded}%
    \renewcommand\color[2][]{}%
  }%
  \providecommand\transparent[1]{%
    \errmessage{(Inkscape) Transparency is used (non-zero) for the text in Inkscape, but the package 'transparent.sty' is not loaded}%
    \renewcommand\transparent[1]{}%
  }%
  \providecommand\rotatebox[2]{#2}%
  \newcommand*\fsize{\dimexpr\f@size pt\relax}%
  \newcommand*\lineheight[1]{\fontsize{\fsize}{#1\fsize}\selectfont}%
  \ifx\svgwidth\undefined%
    \setlength{\unitlength}{226.77165354bp}%
    \ifx\svgscale\undefined%
      \relax%
    \else%
      \setlength{\unitlength}{\unitlength * \real{\svgscale}}%
    \fi%
  \else%
    \setlength{\unitlength}{\svgwidth}%
  \fi%
  \global\let\svgwidth\undefined%
  \global\let\svgscale\undefined%
  \makeatother%
  \begin{picture}(1,0.375)%
    \lineheight{1}%
    \setlength\tabcolsep{0pt}%
    \put(0.07607179,0.25677632){\makebox(0,0)[lt]{\lineheight{1.25}\smash{\begin{tabular}[t]{l}$\upsigma$\end{tabular}}}}%
    \put(0.8709713,0.25677632){\makebox(0,0)[lt]{\lineheight{1.25}\smash{\begin{tabular}[t]{l}$\upsigma$\end{tabular}}}}%
    \put(0,0){\includegraphics[width=\unitlength,page=1]{constitutive-rheological-model-elastoplasticity-ip.pdf}}%
    \put(0.32968738,0.26788635){\makebox(0,0)[t]{\lineheight{1.25}\smash{\begin{tabular}[t]{c}$E$\end{tabular}}}}%
    \put(0.59875298,0.26950123){\makebox(0,0)[t]{\lineheight{1.25}\smash{\begin{tabular}[t]{c}$\upsigma_Y$\end{tabular}}}}%
  \end{picture}%
\endgroup%